\documentclass{article} 
\usepackage{iclr2027_conference,times}

\usepackage{amsmath,amsfonts,bm}

\def\eqref#1{equation~\ref{#1}}

\def\1{\bm{1}}

\DeclareMathAlphabet{\mathsfit}{\encodingdefault}{\sfdefault}{m}{sl}
\SetMathAlphabet{\mathsfit}{bold}{\encodingdefault}{\sfdefault}{bx}{n}

\usepackage{graphicx}
\usepackage{subcaption}
\usepackage{booktabs}
\usepackage{multirow}
\usepackage{makecell}
\usepackage{float}
\usepackage{url}
\usepackage[most]{tcolorbox}
\usepackage{algorithm}
\usepackage{algorithmic}
\usepackage{multirow}
\usepackage[table]{xcolor}
\usepackage{tabularx}
\usepackage{wrapfig}

\usepackage[table]{xcolor}
\definecolor{modelcyan}{HTML}{E0F7FA} 
\definecolor{modelmint}{HTML}{E8F5E9} 
\definecolor{oursblue}{HTML}{EAF0F8}
\definecolor{ourswarm}{HTML}{F7EFE7}
\definecolor{oursgreen}{HTML}{EDF6F2}
\definecolor{cvprblue}{rgb}{0.21,0.49,0.74}

\usepackage[pagebackref,breaklinks,colorlinks,citecolor=cvprblue]{hyperref}

\definecolor{AccentBlue}{RGB}{222, 226, 229}

\let\titleold\title
\renewcommand{\title}[1]{\titleold{#1}\newcommand{\thetitle}{#1}}
\def\maketitlesupplementary
{
    \newpage
    \begin{center}
        \Large
        \textbf{\thetitle}\\
        \vspace{0.5em}Supplementary Material \\
        \vspace{1.0em}
    \end{center}
}
\title{Devils in Question Relay: Source-Conditioned Relay Steering to Mitigate Hallucinations in Audio-visual large language models}

\author{Yu Zhang$^{1,2}$, Pingrui Zhang$^{3}$, Xuefeng Bai$^{1}$, Pengfei Zhang$^{2}$, Yang Xiang$^{2}$, Kehai Chen$^{1,2}$ \\
\textsuperscript{1}Harbin Institute of Technology, Shenzhen, China \\
\textsuperscript{2}Peng Cheng Laboratory, Shenzhen, China \quad
\textsuperscript{3}Fudan University \\
    \texttt{yuzhang2717@gmail.com, \{baixuefeng,chenkehai\}@hit.edu.cn} \\
}

\iclrfinalcopy 
\begin{document}

\maketitle
\lhead{Under review as a conference paper at ICLR 2027}
\begin{abstract}

Audio-visual large language models (AVLLMs) have made
remarkable progress in multimodal understanding and reasoning
through interactions among visual, auditory, and linguistic
information.
However, recent studies show that AVLLMs face a critical
challenge: \textbf{source-confused grounding hallucination}, where cues from the unused modality induce responses that the required modality does not support, undermining
reliability in real-world applications.
Existing methods have made progress in mitigating this failure,
yet how it arises from internal cross-modal interactions
remains insufficiently understood.
To address this gap, we conduct path-intervention and
representation analyses, revealing  
a \emph{question-relay} mechanism: question states carry
interfering cues alongside required-source evidence,
undermining grounding in required-modality evidence.
Cutting pathways from
interfering modality to question states yields greater
correct-answer logit recovery than cutting those to
the generation position.
Motivated by these findings, we propose \textbf{\textsc{Secret}}
(\textbf{S}ourc\textbf{E}-\textbf{C}onditioned \textbf{RE}lay s\textbf{T}eering), a training-free method
that mitigates cross-modal interference at the question relay.
Using contrasting question representations elicited through
different modality-pathway interventions, \textsc{Secret}
steers the original question states toward required-source evidence.
Experiments on two widely adopted benchmarks CMM and AVHBench across three AVLLMs show that \textsc{Secret} consistently outperforms prior training-free methods, substantially mitigating
source-confused grounding hallucinations (e.g., up to +18.0 and +7.1 percentage points over base models).
Modality-specific captioning further demonstrates its
generalizability to open-ended generation.
\end{abstract}
\section{Introduction}

Multimodal large language models (MLLMs)~\citep{qwen25vl,gpt4,gemini2023}
are advancing machine perception toward integrated understanding
of visual, auditory, and textual information.
Recent progress in audio-visual large language models 
(AVLLMs)~\citep{xu2025qwen2,cui2026minicpmo45realtimefullduplex,damonlpsg2024videollama2,Qwen3-Omni}
has demonstrated strong capabilities in multimodal perception,
reasoning, and instruction following.
By combining complementary sensory cues with language
instructions, AVLLMs support richer 
understanding of
complex multimodal inputs, enabling more diverse real-world applications, such as autonomous driving~\citep{zhao2025survey} and human--computer interaction~\citep{gonzalez2026visualaccess}.

\begin{figure}[t]
    \centering
    \includegraphics[width=0.96\textwidth]{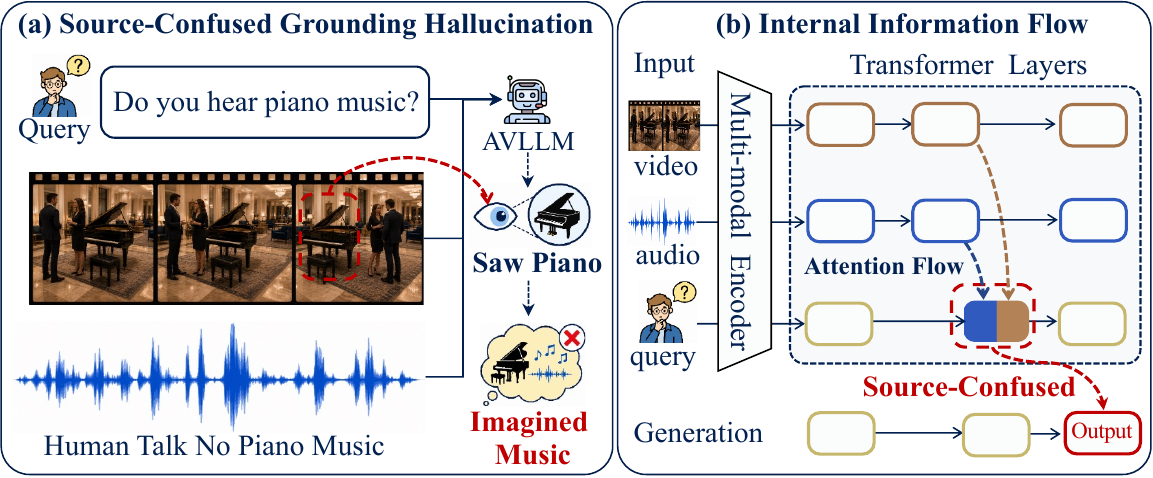}
\caption{
\textbf{(a) Source-confused grounding hallucination in the
audio-required setting.}
The audio contains human speech but no piano music, yet
the visible piano leads the model to hallucinate piano music.
\textbf{(b) Question-relay mechanism.}
Question states relay interfering cues alongside
required-source evidence, allowing non-required information
to influence source-specific answers.
}

    \label{fig:motivation}
\end{figure}

However, recent studies reveal a critical challenge in AVLLMs:
\textbf{source-confused grounding hallucination}, where cues
from a non-required modality induce responses unsupported by
the required modality~\citep{sung2024avhbench,leng2024curse}.
As shown in Fig.~\ref{fig:motivation}(a), a visible piano can
lead the model to hallucinate piano music even when the audio
contains only human speech.
Such failure undermines the reliability of AVLLMs in real-world
applications involving complex audio-visual inputs.
Existing methods mitigate this failure through inference-time
corrective decoding~\citep{chung2026mad,jung2026avcd} or
training-time alignment~\citep{chaubey2026mod,chen2026omnidpo},
yet how it arises from internal cross-modal interactions
remains insufficiently understood.

To investigate this failure, we first find that  source-confused grounding hallucination is not simply due to misunderstanding the requested evidence source.
This motivates a more specific question: \emph{Through which internal pathways do interfering cues influence source-specific answers, and where can this interference be corrected?}
To answer this question, we conduct path-intervention and
representation analyses.
We find that question states, the hidden representations at question-token positions, carry modality information
for subsequent answer prediction, a role we term the
\emph{question relay}.
Specifically, cutting 
pathways from required-modality  to
question states reduces correct-answer support on
source-faithful cases.
Cutting attention pathways from the interfering modality
to question states can partially restore correct-answer support for source-confused cases; and
this intervention yields greater
correct-answer logit recovery at question positions than
at the generation position, 
a focus of prior attention analyses
and interventions~\citep{selvakumar2026really,yu2026causally}.
Together, these findings reveal a \emph{question-relay} mechanism
of source-confused grounding hallucination: interfering cues enter
question states alongside required-source evidence and
influence source-specific answers, as summarized in
Fig.~\ref{fig:motivation}(b).

Motivated by these findings, we propose \textbf{\textsc{Secret}}
(\textbf{S}ourc\textbf{E}-\textbf{C}onditioned
\textbf{RE}lay s\textbf{T}eering), a training-free method
that steers question representations toward required-modality
evidence.
Specifically, \textsc{Secret} constructs source-conditioned
positive and negative question representations by cutting
interfering- and required-modality pathways into question
states, respectively, while retaining the complete
audio-visual input.
It then steers the original question states using the
norm-matched, token-wise difference between these
representations.
\textsc{Secret} substantially mitigates source-confused
grounding hallucinations, improving average accuracy over
the base models by up to 18.0 and 7.1 percentage points on CMM and
AVHBench, and consistently outperforming the evaluated training-free methods across three AVLLMs.
Modality-specific captioning under mismatched audio-video
inputs also demonstrates \textsc{Secret}'s generalizability to open-ended
generation, with lower distractor-reference overlap and higher
modality-grounding scores.
Intervention comparisons and fine-grained behavior analysis
provide a deep understanding of \textsc{Secret}'s effectiveness.

Our contributions are threefold:
(i) we identify a question-relay mechanism of source-confused
grounding hallucinations;
(ii) we propose \textsc{Secret}, a training-free method
for source-conditioned question steering;
and (iii) we demonstrate the effectiveness of the proposed \textsc{Secret} across three
AVLLMs and generalization to modality-specific captioning.

\section{Understanding Source-Confused Grounding}

In this section, we investigate source-confused grounding
hallucination through three progressive analyses.
First, we find that this failure is not simply due to misunderstanding the requested evidence source in question(\S\ref{sec:intent}).
We therefore examine multi-modal information flow during inference,
identifying question states as a relay for required-source
evidence (\S\ref{sec:routing}).
Interfering cues also enter this relay, and cutting their
attention pathways to question states restores correct-answer
support more effectively than cutting those to the
generation position (\S\ref{sec:object}).
These findings motivate source-conditioned relay steering
at the question relay to mitigate cross-modal interference
(\S\ref{sec:method}).

\subsection{Preliminaries}
For an AVLLM $f_\theta$, we abstract input encoding,
projection, and tokenization as a multimodal encoder
that maps video $V$, audio $A$, and question $Q$ to
the token sequence $X=[X_V;X_A;X_Q]$.\footnote{We omit
system and special tokens and group tokens by modality
for notational simplicity; audio and video tokens may
be interleaved in practice~\citep{xu2025qwen2}.}
These tokens are then processed by the LLM backbone,
where we focus our analysis on cross-modal information flow.
We study source-confused grounding hallucination using the \emph{Video-Driven Audio Hallucination}
and \emph{Audio-Driven Video Hallucination} subsets of
AVHBench~\citep{sung2024avhbench}, where each textual question explicitly specifies whether the answer should be grounded in audio or video evidence.
Let $r\in\{A,V\}$ denote the required modality and
$\bar r$ the other modality.
A \emph{source-faithful} answer is supported by evidence
from $r$, whereas a \emph{source-confused} prediction
incorrectly relies on cues from $\bar r$, as illustrated
in Fig.~\ref{fig:motivation}(a).
Dataset details and statistics are provided in
Apdx~\ref{supp:data}.

\subsection{\textsc{Observation~1}: AVLLM Can Reliably Identify the Required Modality}
\label{sec:intent}

We begin with a fundamental diagnostic question: \emph{Can a AVLLM identify which evidence modality the textual question explicitly requires?} 
This test assesses the model's ability to identify the
required evidence source from the textual question.
We provide Qwen2.5-Omni-7B with the textual question and ask it to classify the required evidence as \emph{audio}, \emph{video}, or \emph{ambiguous}, without answering the original question. 
See Apdx.~\ref{supp:classify} for experimental details.
The model correctly identifies the required modality for 99.85\% of the questions, demonstrating that it can reliably recover the source of required evidence from the question alone.
This suggests that source-confused grounding hallucination is not simply due to a failure to identify the required modality.

\noindent\textbf{Takeaways.}
These results suggest that source-confused grounding
is not simply due to misunderstanding which modality
the question requires.
Therefore we next investigate how required-source evidence
and interfering cues are routed within the model
and influence its predictions (\S\ref{sec:routing}, \S\ref{sec:object}).

\subsection{\textsc{Observation~2}: Question States Relay Required-Source Evidence}
\label{sec:routing}
Following \textsc{Observation~1}, we first examine how evidence from the required modality is routed through the model to support source-faithful predictions in this section.

\noindent\textbf{Method.}
We use attention-path cutting~\citep{cross_modal} analysis on source-faithful cases for Qwen2.5-Omni-7B to identify 
the critical pathway that supports the model's prediction
At layer $\ell$, the attention output for target token $t$ is computed through multi-head self-attention:
\begin{equation}
\mathbf{A}_{t}^{\ell}
=
\sum_{j=1}^{J}
\operatorname{Softmax}
\left(
\frac{
\mathbf{q}_{t}^{\ell,j}
(\mathbf{K}^{\ell,j})^{\top}
}{
\sqrt{d}
}
+
\mathbf{M}_{t,:}^{\ell}
\right)
\mathbf{V}^{\ell,j}
\mathbf{W}_{O}^{\ell,j}.
\end{equation}
Here $J$ is the number of heads and $d$ is the per-head query dimension.
For head $j$, $\mathbf q_t^{\ell,j}$ is the target query, $\mathbf K^{\ell,j}$ and $\mathbf V^{\ell,j}$ are the key and value matrices, $\mathbf W_O^{\ell,j}$ is the output projection and $\mathbf{M}_{t,:}^{\ell}$ denotes the causal-mask row for target token $t$.
For a source token set $S$ and a target token set $T$, we define the  attention pathway as
$\mathcal{P}_{S\rightarrow T}
=
\left\{
(s,t)\mid s\in S,\ t\in T
\right\}$,
where $(s,t)$ denotes an attention edge through which target token $t$ attends to source token $s$. 
We cut this pathway across a seven-layer window
$\mathcal{W}_{\ell}$ centered on layer $\ell$ by modifying
the corresponding attention-mask entries:
\begin{equation}
\widetilde{\mathbf{M}}_{t,s}^{m}
=
\begin{cases}
-\infty,
& (s,t)\in\mathcal{P}_{S\rightarrow T}
  \ \text{and}\ m\in\mathcal{W}_{\ell},\\
\mathbf{M}_{t,s}^{m},
& \text{otherwise}.
\end{cases}
\label{eq:path_mask}
\end{equation}
Other mask entries retain their original values and token groups denote sets of token positions.

\noindent\textbf{Metric.}
We measure the effect of cutting each attention pathway
using the mean relative change in target-answer probability.
More negative values indicate a larger reduction in
target-answer probability, suggesting that the model
relies more strongly on the pathway for prediction.
See Apdx~\ref{supp:route_cutting} for more details
and robustness analysis across different AVLLMs and cutting
window sizes.

\noindent\textbf{Results.}
Let $X_G$ denote the final position of the complete tokenized
prompt, where the model predicts the first answer token.
We call this the \emph{generation position} and exclude it
from the question-token positions $X_Q$.
The source set $S$ consists of the tokens of instruction-required modality, $X_r$, while the destination set $T$ is chosen from $X_Q$, the tokens of interfering modality $X_{\bar r}$, and $X_G$.
We analyze examples with 
source-faithful predictions
in the Video-Driven
\begin{wrapfigure}{r}{.62\textwidth}
\vspace{-2mm}
\centering

\includegraphics[width=0.495\linewidth]{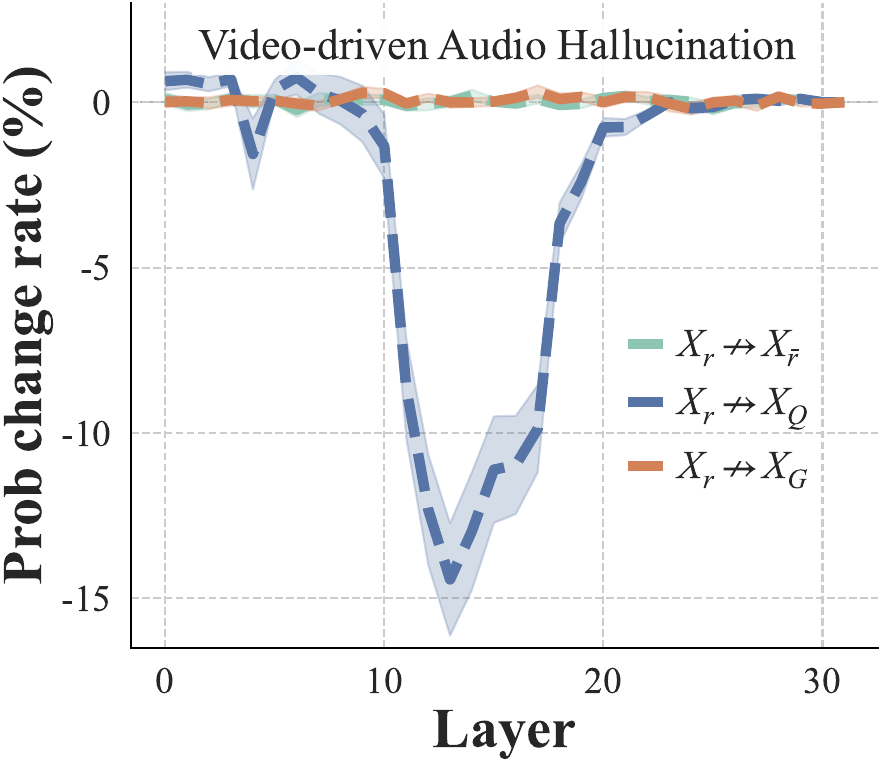}%
\hfill
\includegraphics[width=0.495\linewidth]{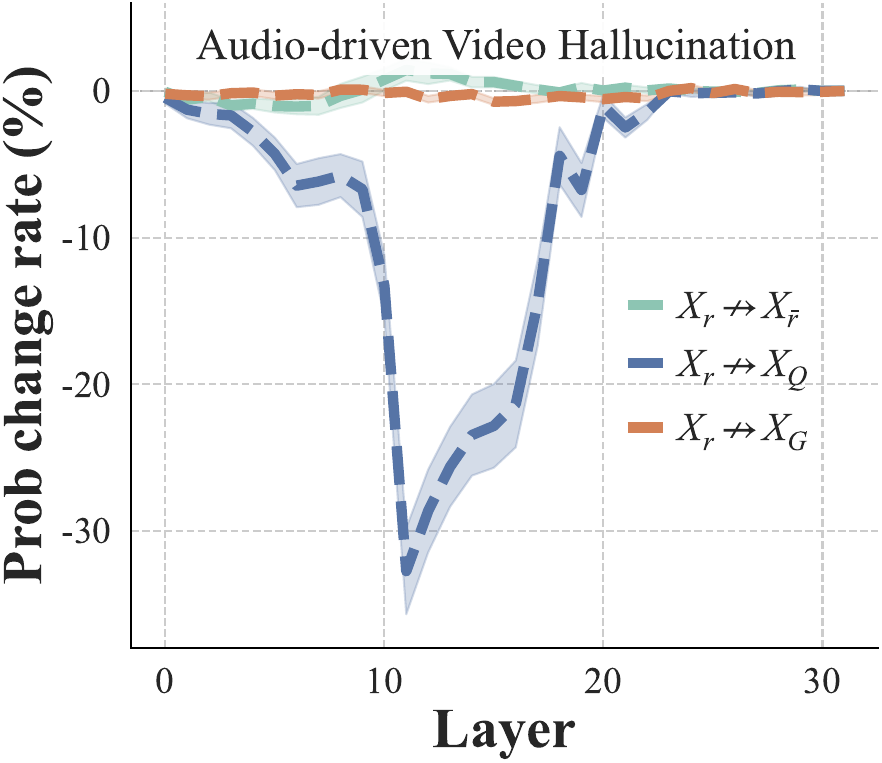}
        \vspace{-2pt}
\caption{Layer-wise effects of cutting attention pathways from the tokens of required modality $X_r$ to the tokens of interfering modality $X_{\bar r}$, question tokens $X_Q$ and the generation position $X_G$ for source-faithful prediction.
Cutting the attention pathway from the tokens of required-modality to question tokens produces the largest reduction, suggesting that the pathway is the most important for source-faithful prediction.}
\label{fig:finding1_2}
\vspace{-4mm}
\end{wrapfigure}
Audio Hallucination and Audio-Driven
Video Hallucination settings, where the required modalities
are audio and video, respectively.
Across both settings, Fig.~\ref{fig:finding1_2} shows that
cutting $X_r\!\rightarrow\!X_Q$ produces the largest
decrease in target-answer probability among the three
interventions.
This suggests that source-faithful predictions rely more
strongly on $X_r\!\rightarrow\!X_Q$ than on
$X_r\!\rightarrow\!X_G$ or
$X_r\!\rightarrow\!X_{\bar r}$.
While prior work~\citep{selvakumar2026really} examines
audio-visual evidence use at generation positions,
our analysis highlights question states as an intermediate
relay carrying required-source evidence to answer prediction,
a role we term the \textbf{question relay}.
We next examine whether non-required cues enter this relay and are mistaken for required-source evidence
(\S\ref{sec:object}).

\noindent\textbf{Takeaways.}
Source-faithful predictions
depend most strongly on the pathway from required-modality
 to question tokens, highlighting question states
as a key relay for required-source evidence.

\subsection{\textsc{Observation~3}: Interfering Cues in Question States Influence Predictions}
\label{sec:object}

\textsc{Observation~2} identifies question state as a relay for required-source evidence. We next examine whether cues from interfering modality enter this relay and lead to source-confused hallucination.

\noindent\textbf{Cutting the interfering route attenuates wrong-source evidence.}
We probe interfering information in question states
by measuring their support for the target object associated
with the hallucinated answer.
An LLM parser extracts the target object from the question, such as the ``piano" in Fig.~\ref{fig:motivation}(a).
We use Logit Lens~\citep{geva2022transformer} to measure its layer-wise \textit{target-object score} within $X_Q$. 
A higher score indicates stronger support for the object
in the question states.
We compare the original run (\textit{Original}) with a run
that cuts $X_{\bar r}\!\rightarrow\!X_Q$
(\textit{Intervened}), keeping the inputs unchanged.
Following \textsc{Observation~2}, we focus on Layers 10--20,
where modality-to-question interventions have the strongest
effects.
Object extraction and score computation are detailed in
Apdx~\ref{supp:object_evidence}.

\begin{figure}[t]
    \centering
    \begin{subfigure}[t]{0.66\textwidth}
        \centering
        \includegraphics[width=0.495\linewidth]{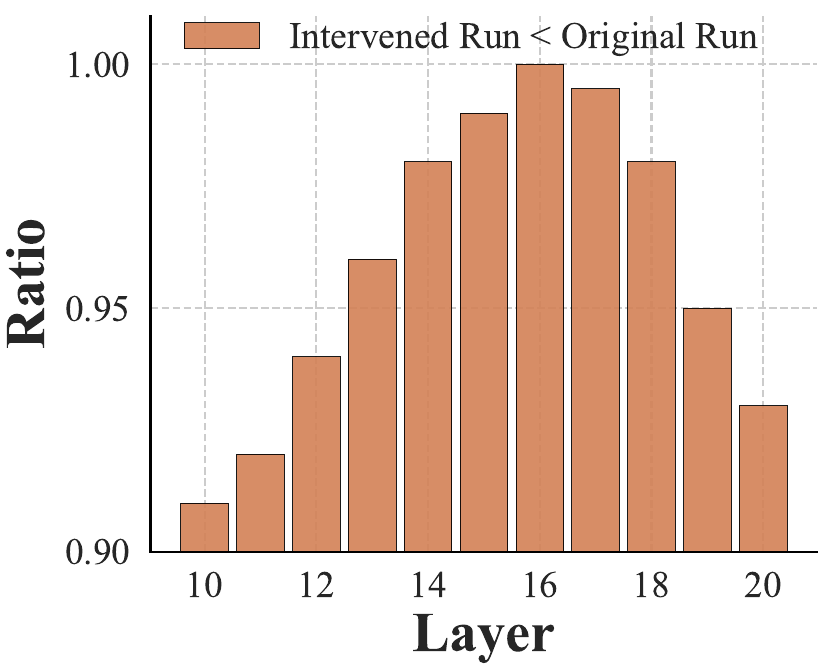}%
        \hfill
        \includegraphics[width=0.495\linewidth]{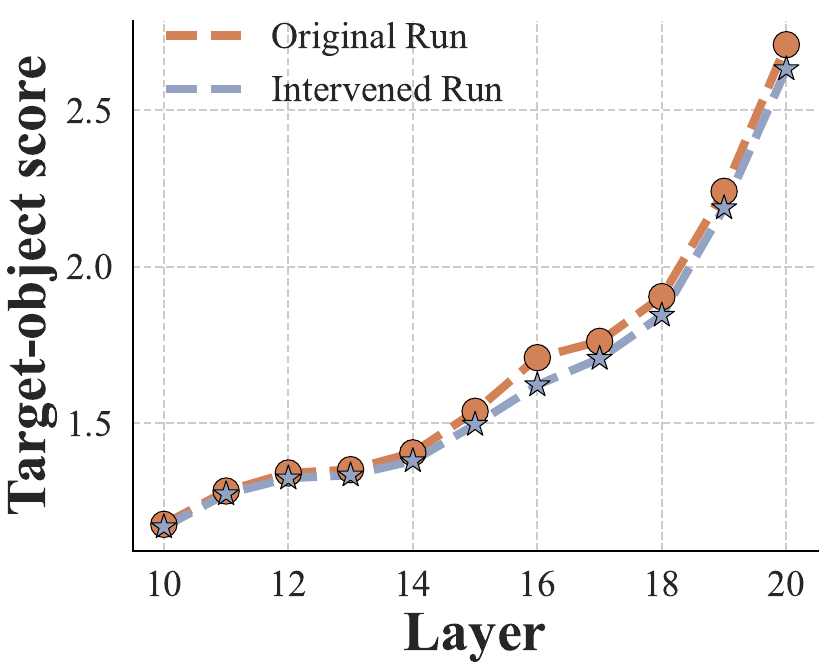}
        \caption{Target-object score comparison in question states.}
    \end{subfigure}%
    \hfill
    \begin{subfigure}[t]{0.33\textwidth}
        \centering
        \includegraphics[width=\linewidth]{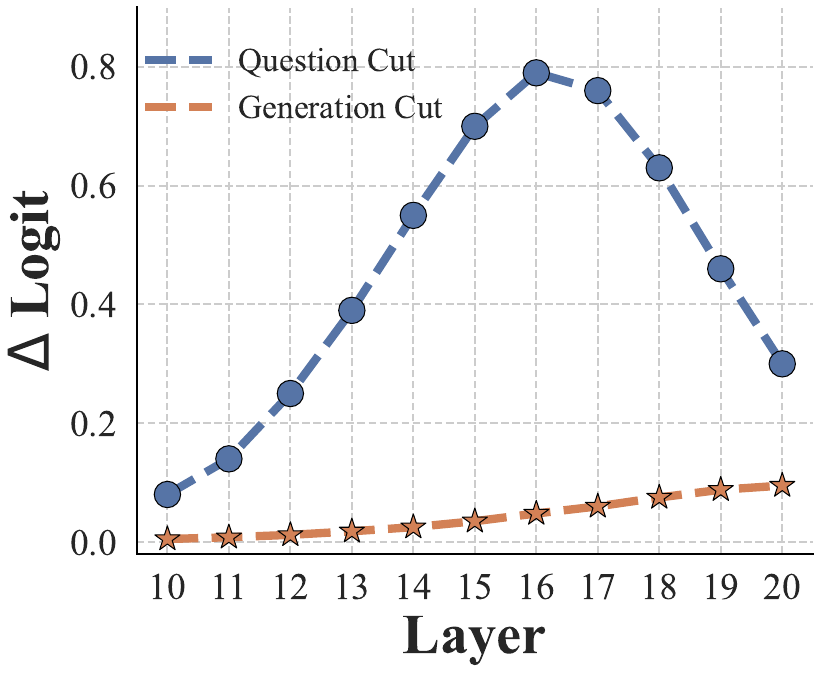}
        \caption{Logit recovery.}
    \end{subfigure}
    \vspace{-5pt}

    \caption{
    \textbf{Interfering cues in question states and their effects on hallucination predictions.}
    \textbf{(a)} \textit{Original} is the unmodified run;
    \textit{Intervened} cuts pathways from interfering-modality 
    into question states.
    \textbf{Left:} layer-wise fraction of examples with lower \textit{target-object
    scores} in \textit{Intervened} than in \textit{Original},
    exceeding 90\% at every tested layer.
    \textbf{Right:} mean layer-wise \textit{target-object scores} across all samples, showing lower scores after intervention.
    \textbf{(b)} \textit{Question Cut} and \textit{Generation Cut} cut pathways from interfering-modality to question states and the generation position, respectively.
    For each intervention, $\Delta$Logit is the correct-answer
    output logit at $X_G$ after intervention minus that in
    \textit{Original}.
    \textit{Question Cut} yields greater recovery over most
    tested layers.
    }
    \label{fig:finding3}
    \vspace{-10pt}
\end{figure}

Fig.~\ref{fig:finding3}(a) shows that the fraction of examples
with lower target-object scores in \textit{Intervened} than
in \textit{Original} exceeds 90\% at every tested layer,
approaching 100\% at several layers (left).
The mean target-object score across examples is also lower
in \textit{Intervened} than in \textit{Original}, indicating
weaker target-object signals in question states after
cutting the interfering pathway (right).
Together, these results suggest that the interfering-modality
pathway carries wrong-source cues into question states.

\noindent\textbf{Question cut yields greater correct-answer
logit recovery.}
Having identified interfering signals in question states,
we next compare question tokens and the generation position
as intervention targets for restoring correct-answer support.
Specifically, we compare cutting
$X_{\bar r}\!\rightarrow\!X_Q$ (\textit{Question Cut})
with cutting $X_{\bar r}\!\rightarrow\!X_G$
(\textit{Generation Cut}).
The latter position has been a focus of prior analyses
and interventions~\citep{selvakumar2026really,yu2026causally}.
We compare the effects over Layers 10--20 using
\emph{Correct-Answer Logit Recovery} ($\Delta$Logit):
the correct-answer logit at $X_G$ after intervention
minus that in \textit{Original}.
As shown in Fig.~\ref{fig:finding3}(b),
\textit{Question Cut} yields markedly larger $\Delta$Logit
than \textit{Generation Cut} at every tested layer,
indicating more effective recovery of correct-answer support.

\noindent\textbf{Takeaways.}
\textsc{Observation~2 \& 3} identify
question states as a relay for both required-source evidence
and interfering cues.
Attention interventions at question tokens restore
correct-answer support more effectively than those
at the generation position commonly targeted in prior work.

\section{Source-Conditioned Relay Steering (\textsc{Secret})}
\label{sec:method}
\begin{figure}[t]
    \centering
    \includegraphics[width=0.95\textwidth]{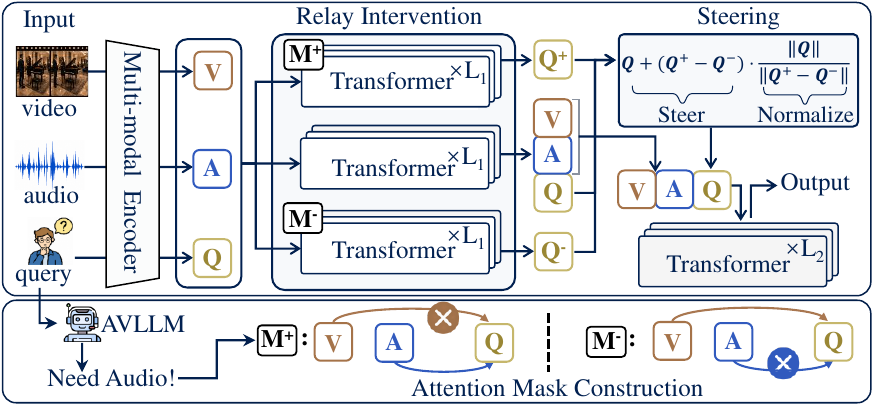}
\caption{
\textbf{Overview of the proposed \textsc{Secret}.}
\textbf{Bottom:} The AVLLM predicts the required modality from the question alone to construct $\mathbf{M}^{+}$ and $\mathbf{M}^{-}$, which cut attention from the
interfering and required modalities to question tokens, respectively.
\textbf{Top:} Three parallel branches with shared parameters process
the same input through the first $L_1$ layers, yielding
positive, original, and negative question states.
Each token-wise positive--negative difference is scaled
to the original state's L2 norm and added to it.
Updated question states and original audio-visual states
pass through the remaining $L_2$ layers for generation.
}
\vspace{-10pt}
    
    \label{fig:method}
\end{figure}

Our analyses show that question states relay both
required-source evidence and interfering cues, and that
intervening at this relay can effectively restore correct-answer support.
Building on this finding, we propose \textsc{Secret} (\textbf{S}ourc\textbf{E}-\textbf{C}onditioned \textbf{RE}lay s\textbf{T}eering), as illustrated in Fig.~\ref{fig:method}.
We elicit source-conditioned question representations by selectively cutting modality-to-question attention pathways.
Guided by these representations, we steer the original question states to favor required-source evidence over interfering-modality cues.

\subsection{Required-Modality Identification}

Following \textsc{Observation~1} (\S\ref{sec:intent}),
we prompt the AVLLM $f_\theta$ with the textual question
alone to predict the required modality $\hat r\in\{A,V\}$.
This prediction guides the construction of two attention
masks for eliciting positive and negative question
representations.
Both masks follow Eq.~(\ref{eq:path_mask}) and target
question-token positions ($T=X_Q$), excluding the generation position $X_G$.
The positive mask $\mathbf{M}^{+}$ cuts pathways from
$S=X_{\bar{\hat r}}$ to $T=X_Q$, where $\bar{\hat r}$ denotes the
other modality, while preserving the required-modality
pathway.
Conversely, the negative mask $\mathbf{M}^{-}$ cuts
pathways from $S=X_{\hat r}$ to $T=X_Q$ while preserving the
interfering-modality pathway.
The resulting representations provide positive and
negative references for subsequent question-state steering.

\subsection{Question-Relay Steering}
\label{sec:method_relay_steer}
We construct positive and negative question representations through pathway interventions to steer the original states toward required-source evidence.

\noindent\textbf{Relay intervention.}
We process the encoded input $X=[X_V;X_A;X_Q]$ through
three parallel branches sharing the parameters of the
first $L_1$ Transformer layers.
Throughout these layers, the positive branch applies
$\mathbf{M}^{+}$ to block attention from the interfering
modality to question tokens, while the negative branch
applies $\mathbf{M}^{-}$ to block attention from the
required modality.
The original branch retains the unmodified attention mask.
At layer $L_1$, these branches yield
$\mathbf{H}_Q^{+},\mathbf{H}_Q^{-},\mathbf{H}_Q
\in\mathbb{R}^{|X_Q|\times d_h}$, respectively,
where $|X_Q|$ is the number of question tokens and $d_h$
is the hidden dimension.
We omit layer superscripts for clarity.
As in our attention-routing analysis, $X_Q$ excludes
the generation position $X_G$.
Corresponding rows across the three matrices represent
the same question token.

\noindent\textbf{Source-Conditioned Question Steering.}
For each question position $i\in X_Q$, let
$\mathbf{h}_i^{+}$, $\mathbf{h}_i^{-}$, and $\mathbf{h}_i$
denote the positive, negative, and original states, respectively.
We use their token-wise difference,
$\Delta\mathbf{h}_i=\mathbf{h}_i^{+}-\mathbf{h}_i^{-}$,
to steer the original state toward required-source evidence.
To balance correction strength and generation
stability~\citep{liu2023context,zou2023representation,zhang2025modalitypreference},
we scale each direction to match the original state's
L2 norm before adding it:
\begin{equation}
\widetilde{\mathbf{h}}_i
=
\mathbf{h}_i+
\frac{\|\mathbf{h}_i\|_2}
     {\|\Delta\mathbf{h}_i\|_2}
\Delta\mathbf{h}_i,
\quad i\in X_Q.
\label{eq:question_steering}
\end{equation}
We combine the steered question states
$\widetilde{\mathbf{H}}_Q$ with the original branch's
audio and video states in their original token order.
The sequence then passes through the remaining $L_2$
Transformer layers to generate the answer.
Steering is applied only during prefill.
The remaining layers cache the keys and values derived
from the updated sequence for subsequent autoregressive
generation.
See Apdx~\ref{supp:secret_implementation} for implementation details of \textsc{Secret}.

\noindent\textbf{Distinction from prior work.}
Motivated by the relay role and stronger intervention effects at question
tokens (\S\ref{sec:routing}, \S\ref{sec:object}),
\textsc{Secret} targets modality-to-question pathways
rather than the modality-to-generation pathways commonly
used in prior work~\citep{yu2026causally,zhou2025mitigating}.
Unlike methods that intervene by perturbing or removing
modality inputs~\citep{chung2026mad,jung2026avcd},
\textsc{Secret} constructs contrasting representations
through internal attention-path interventions,
preserving audio-visual context and avoiding
potential representational shifts from altered inputs.
RQ1 (\S\ref{sec:analysis}) compares these intervention methods
to assess the benefits of targeting question states.

\section{Experiments}
\begin{table*}[t]
    \centering
      \renewcommand{\arraystretch}{1.0}
      \setlength{\tabcolsep}{4.5pt}
      \caption{Main results on CMM and AVHBench in accuracy (\%).
      Overall Acc. denotes the mean of the two subset accuracies for each benchmark.
      Values in parentheses indicate absolute gains in percentage points over the corresponding base model.}
      \vspace{-5pt}
      \resizebox{1.0\linewidth}{!}{
      \begin{tabular}{lcccccc}
      \toprule
      \multirow{3}{*}[-0.8ex]{\textbf{Method}} &
      \multicolumn{3}{c}{\textbf{CMM}} &
      \multicolumn{3}{c}{\textbf{AVHBench}} \\
      \cmidrule(lr){2-4}
      \cmidrule(lr){5-7}
       & \multirow{2}{*}{Visual Dom.} &
       \multirow{2}{*}{Audio Dom.} &
       \multirow{2}{*}{\textbf{Overall Acc.}} &
       Video-Driven & Audio-Driven &
       \multirow{2}{*}{\textbf{Overall Acc.}} \\
       & & & & Audio Hall. & Video Hall. & \\
      \midrule

      VideoLLaMA2-AV-7B
      & 71.8 & 80.0 & 75.9 & 75.7 & 79.0 & 77.4 \\
      \quad+ VCD~{\scriptsize\textcolor{gray}{(CVPR'24)}}
      & 71.3 & 83.3 & 77.3 & 66.0 & 74.8 & 70.4 \\
      \quad+ AVCD~{\scriptsize\textcolor{gray}{(NeurIPS'25)}}
      & 71.8 & 84.0 & 77.9 & 78.3 & 80.3 & 79.3 \\
      \quad+ MAD~{\scriptsize\textcolor{gray}{(CVPR'26)}}
      & 82.3 & 84.3 & 83.3 & 79.7 & 79.1 & 79.4 \\
      \rowcolor{ourswarm}
      \quad+ \textbf{\textsc{Secret}}
      & \textbf{87.3}\,{\scriptsize(+15.5)}
      & \textbf{91.3}\,{\scriptsize(+11.3)}
      & \textbf{89.3}\,{\scriptsize(+13.4)}
      & \textbf{80.6}\,{\scriptsize(+4.9)}
      & \textbf{81.3}\,{\scriptsize(+2.3)}
      & \textbf{81.0}\,{\scriptsize(+3.6)} \\
      \midrule

      Qwen2.5-Omni-7B
      & 64.5 & 72.3 & 68.4 & 73.0 & 80.7 & 76.9 \\
      \quad+ VCD~{\scriptsize\textcolor{gray}{(CVPR'24)}}
      & 62.5 & 71.3 & 66.9 & 70.3 & 77.1 & 73.7 \\
      \quad+ AVCD~{\scriptsize\textcolor{gray}{(NeurIPS'25)}}
      & 66.3 & 72.8 & 69.5 & 75.8 & 79.7 & 77.8 \\
      \quad+ MAD~{\scriptsize\textcolor{gray}{(CVPR'26)}}
      & 76.8 & 84.3 & 80.5 & 78.7 & 84.4 & 81.6 \\
      \rowcolor{ourswarm}
      \quad+ \textbf{\textsc{Secret}}
      & \textbf{84.8}\,{\scriptsize(+20.3)}
      & \textbf{88.0}\,{\scriptsize(+15.7)}
      & \textbf{86.4}\,{\scriptsize(+18.0)}
      & \textbf{82.7}\,{\scriptsize(+9.7)}
      & \textbf{85.3}\,{\scriptsize(+4.6)}
      & \textbf{84.0}\,{\scriptsize(+7.1)} \\
      \midrule

      Qwen3-Omni-30B-A3B
      & 81.3 & 77.0 & 79.2 & 77.0 & 76.6 & 76.8 \\
      \quad+ MAD~{\scriptsize\textcolor{gray}{(CVPR'26)}}
      & 82.8 & 84.5 & 83.6 
      & 79.6 & 80.6 & 80.1 \\
      \rowcolor{ourswarm}
      \quad+ \textbf{\textsc{Secret}}
      & \textbf{85.6}\,{\scriptsize(+4.3)}
      & \textbf{89.8}\,{\scriptsize(+12.8)}
      & \textbf{87.7}\,{\scriptsize(+8.5)}
      & \textbf{81.1}\,{\scriptsize(+4.1)}
      & \textbf{81.6}\,{\scriptsize(+5.0)}
      & \textbf{81.4}\,{\scriptsize(+4.6)}\\
      \bottomrule
      \end{tabular}
      }
      \vspace{-5pt}
      \label{tab:main_results}
  \end{table*}

\subsection{Experimental Setup}

\noindent\textbf{Benchmarks and Metrics.}
Focusing on source-confused grounding hallucination,
we evaluate \textsc{Secret} on two established cross-modal hallucination benchmarks, AVHBench~\citep{sung2024avhbench} and CMM~\citep{leng2024curse}.
For AVHBench, we use the Video-Driven Audio Hallucination and Audio-Driven Video Hallucination, comprising 3,426 question--answer pairs in total.
For CMM, we use the visual-dominance (Visual Dom.) and audio-dominance (Audio Dom.), comprising 800 questions in total.
We report subset accuracies and their arithmetic mean for each benchmark.

\noindent\textbf{Baselines.}
We evaluate \textsc{Secret} across VideoLLaMA2-AV~\citep{damonlpsg2024videollama2}, Qwen2.5-Omni-7B~\citep{xu2025qwen2}, and Qwen3-Omni-30B-A3B~\citep{Qwen3-Omni}.
We compare against training-free hallucination mitigation methods:
VCD~\citep{leng2024mitigating}, contrasting output logits from full and modality-removed inputs;
AVCD~\citep{jung2026avcd}, constructing perturbed branches by selectively masking high-attention tokens in less dominant modalities;
and MAD~\citep{chung2026mad}, using the AVLLM's self-assessed modality relevance to adaptively balance modality-specific contributions during decoding.
Following MAD, we adopt the four-branch audio-visual extension of VCD, and denote this variant as VCD in the tables. See Apdx~\ref{supp:baselines} for more details of baselines.

\subsection{Main Results}

Table~\ref{tab:main_results} reports results on CMM and
AVHBench across three AVLLMs, spanning different model
scales and both dense and mixture-of-experts architectures.
\textsc{Secret} improves overall accuracy over the base
models by up to 18.0 and 7.1 percentage points on CMM
and AVHBench, respectively.
For each backbone, \textsc{Secret} achieves the highest overall
accuracy among the evaluated methods on both benchmarks
and \textsc{Secret} achieves particularly large gains
on CMM.
These gains further support the cross-dataset
applicability of question-relay steering motivated
by our findings on AVHBench.
The gains vary with the required modality.
VideoLLaMA2-AV-7B and Qwen2.5-Omni-7B obtain larger improvements
on audio-required tasks, whereas Qwen3-Omni-30B-A3B benefits more
on video-required tasks.
This variation may reflect differences in baseline
capabilities and susceptibility to cross-modal
interference across models and tasks.

\subsection{Analysis and Discussion}
\label{sec:analysis}
We organize our analysis around four research questions:
(i) \textbf{RQ1}: How does \textsc{Secret} improve upon existing alternative intervention designs?
(ii) \textbf{RQ2}: Which layers are most effective for question
steering, and why?
(iii) \textbf{RQ3}: Does \textsc{Secret} generalize to open-ended tasks?
(iv) \textbf{RQ4}: How does \textsc{Secret} perform under finer-grained evaluation?

\begin{figure}[t]
    \centering

    \begin{subfigure}[t]{0.325\textwidth}
        \centering
        \includegraphics[width=\linewidth]{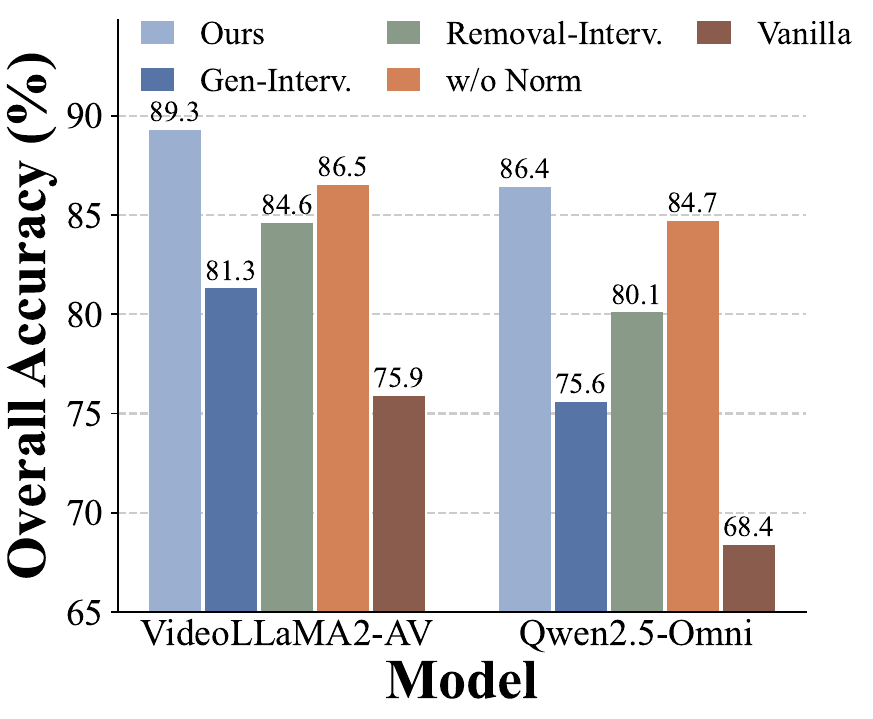}
\caption{Interventions comparison.}

    \end{subfigure}
    \hfill
    \begin{subfigure}[t]{0.325\textwidth}
        \centering
        \includegraphics[width=\linewidth]{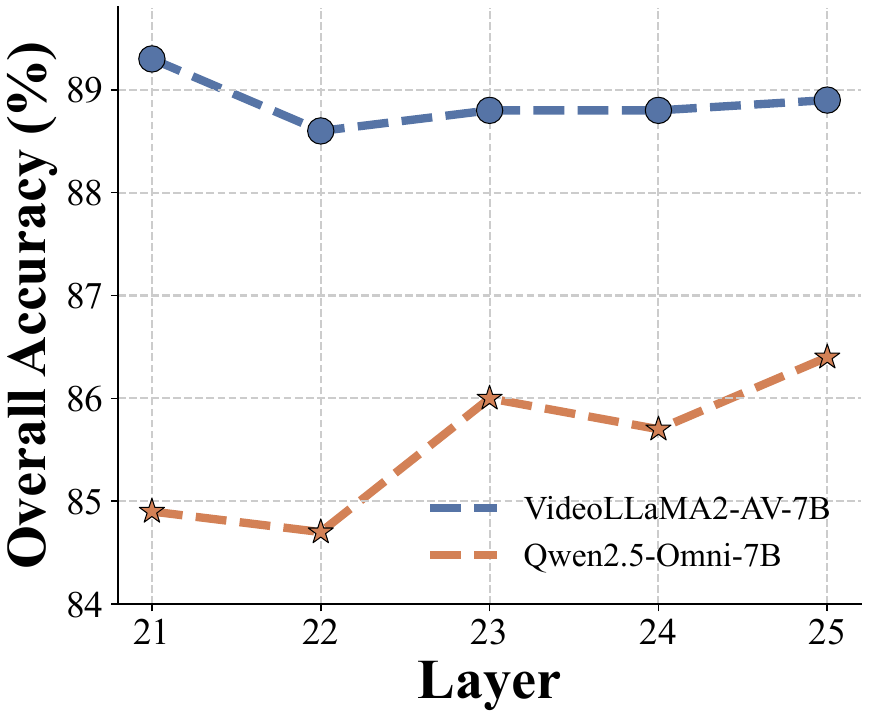}
\caption{Steering depth comparison.}

    \end{subfigure}
    \hfill
    \begin{subfigure}[t]{0.325\textwidth}
        \centering
        \includegraphics[width=\linewidth]{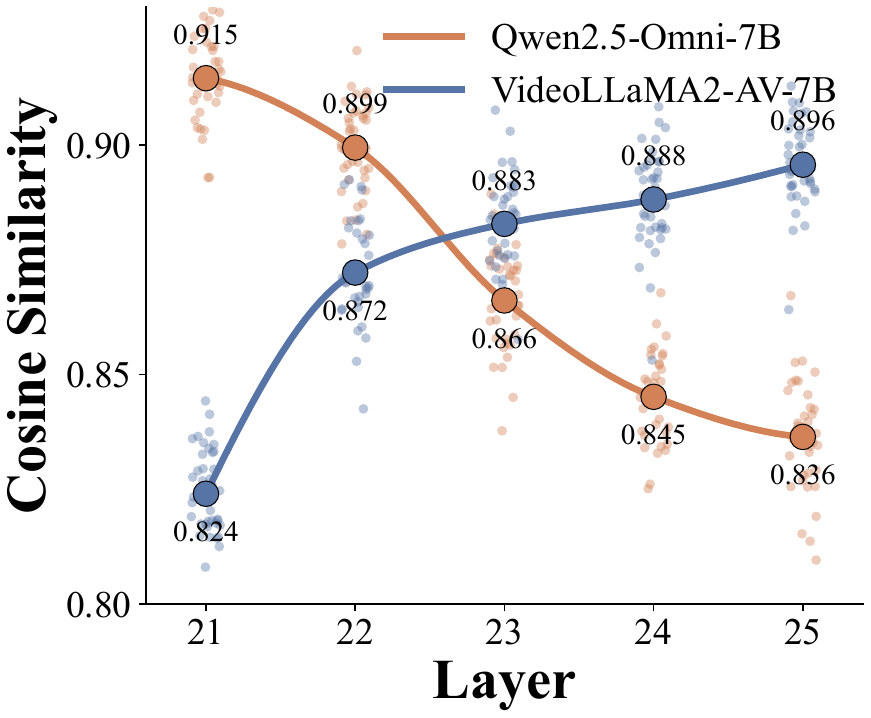}
\caption{Cosine similarity analysis.}
    \end{subfigure}

\vspace{-5pt}
\caption{
\textbf{Ablation and representation analysis on CMM.}
\textbf{(a)} Overall accuracy of \textsc{Secret}, its three
intervention variants, and the original AVLLMs.
\textbf{(b)} Overall accuracy across relay-intervention depths
$L_1$, with question steering applied after the first $L_1$ layers.
\textbf{(c)} Cosine similarity between the positive and negative
question representations at the corresponding depths.
}
    \vspace{-5pt}
\label{fig:ablation}

\end{figure}

\noindent\textbf{RQ1: How does \textsc{Secret} improve upon existing alternative intervention designs?}
Fig.~\ref{fig:ablation}(a) compares \textsc{Secret} with three
variants of its steering framework
(\S\ref{sec:method_relay_steer}) on CMM, using
VideoLLaMA2-AV-7B and Qwen2.5-Omni-7B.
\textit{Gen-Interv.} constructs contrasts through attention
interventions at $X_G$ and applies steering at the same position.
\textit{Removal-Interv.} constructs contrasts from
modality-removed inputs and applies steering at question
positions.
\textit{w/o Norm} omits norm matching in
Eq.~\ref{eq:question_steering}.
See Apdx~\ref{supp:baselines} for comparison settings.
\textsc{Secret} outperforms both \textit{Gen-Interv.} and
\textit{Removal-Interv.} on both models, supporting the
advantage of question-relay steering over alternative
intervention designs adapted from prior work.
Removing norm matching also reduces accuracy on both
models, demonstrating its contribution to steering
performance.

\begin{table*}[t]
    \centering
\caption{Source-grounding evaluation in audio-target and video-target
captioning. T-CIDEr$\uparrow$ measures target-reference agreement;
D-CIDEr$\downarrow$ measures distractor-reference overlap;
LLM-score$\uparrow$ jointly assesses target fidelity and distractor leakage.}
    \label{tab:ablate}
    \vspace{-5pt}

    \renewcommand{\arraystretch}{0.9}
    \setlength{\tabcolsep}{4.5pt}

    \resizebox{0.9\linewidth}{!}{%
    \begin{tabular}{lcccccc}
        \toprule
        \multirow{2}{*}[-0.8ex]{\textbf{Method}} &
        \multicolumn{3}{c}{\textbf{Audio-target Caption}} &
        \multicolumn{3}{c}{\textbf{Video-target Caption}} \\
        \cmidrule(lr){2-4}
        \cmidrule(lr){5-7}

        & T-CIDEr$\uparrow$
        & D-CIDEr$\downarrow$
        & LLM-score$\uparrow$
        & T-CIDEr$\uparrow$
        & D-CIDEr$\downarrow$
        & LLM-score$\uparrow$ \\
        \midrule

        Qwen2.5-Omni-7B
        & \textbf{17.7} & 5.17 & 3.12
        & 29.3 & 3.70 & 3.57 \\

        \quad +Gen-Interv.
        & 17.2 & 5.10 & 3.05
        & 30.2 & 3.10 & 3.79 \\

        \quad +Removal-Interv.
        & 16.5 & 5.43 & 3.55
        & 31.1 & 2.67 & 3.49 \\

        \rowcolor{ourswarm}
        \quad +\textbf{\textsc{Secret}}
        & 17.4
        & \textbf{4.59}
        & \textbf{3.76}
        & \textbf{31.8}
        & \textbf{2.66}
        & \textbf{4.01} \\
        \midrule

        VideoLLaMA2-AV-7B
        & 13.9 & 20.2 & 2.93
        & 32.9 & 5.43 & 3.27 \\

        \quad +Gen-Interv.
        & \textbf{14.8} & 17.6 & 3.18
        & \textbf{34.1} & 4.60 & 3.68 \\

         \quad +Removal-Interv.
        & 12.7 & 15.4 & 2.88
        & 31.6 & 4.10 & 3.46 \\

        \rowcolor{ourswarm}
        \quad +\textbf{\textsc{Secret}}
        & 14.4
        & \textbf{12.8}
        & \textbf{3.57}
        & 32.4
        & \textbf{3.20}
        & \textbf{3.79} \\

        \bottomrule
    \end{tabular}%
    }

    \vspace{-5pt}
\end{table*}

\begin{figure}[t]
    \centering
    \begin{subfigure}[t]{0.33\textwidth}
        \centering
        \includegraphics[width=0.99\linewidth]{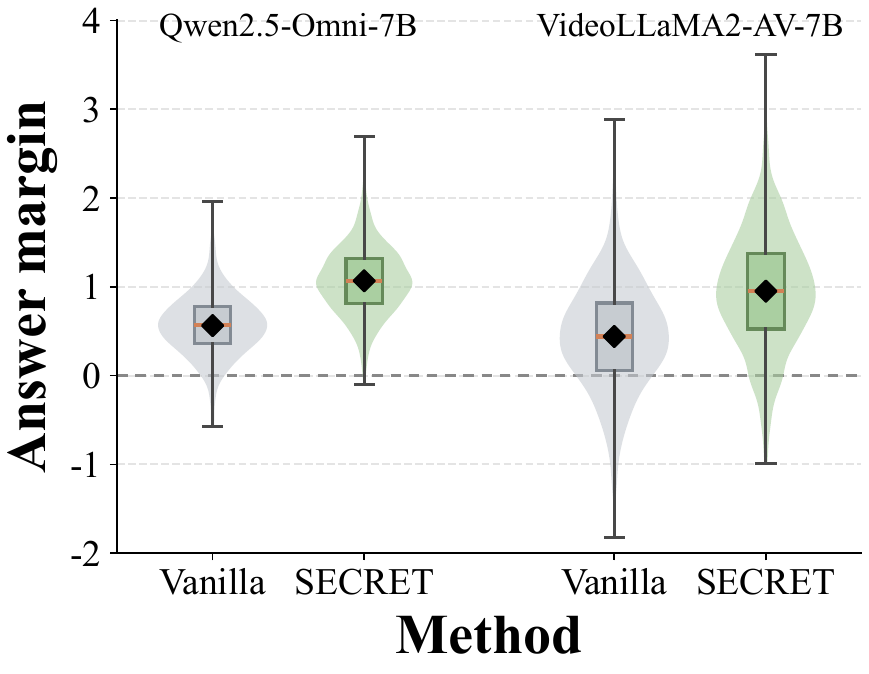}
       \caption{Answer margin distribution.}
    \end{subfigure}
    \hfill
    \begin{subfigure}[t]{0.66\textwidth}
        \centering
        \includegraphics[width=0.495\linewidth]{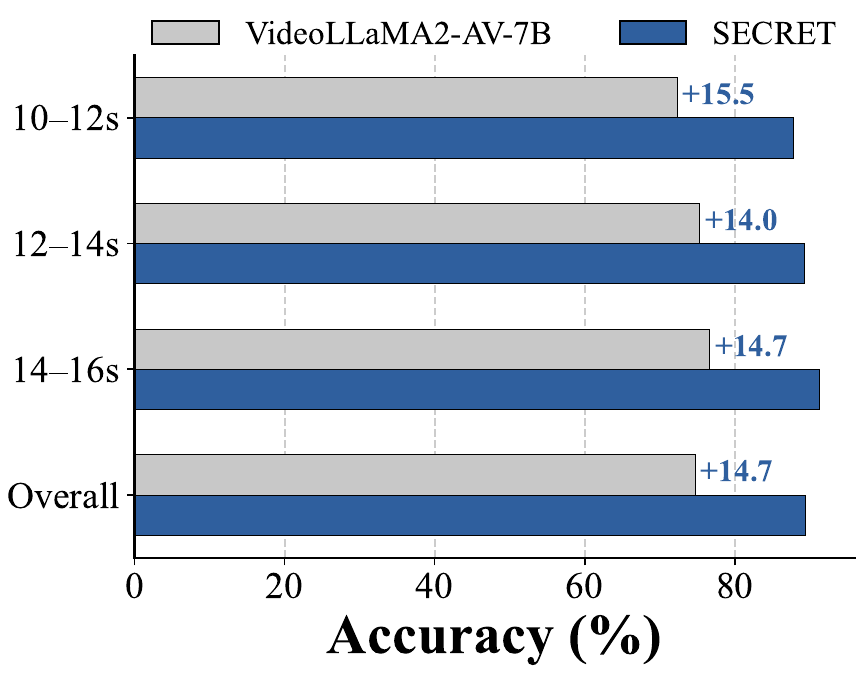}%
        \hfill
        \includegraphics[width=0.495\linewidth]{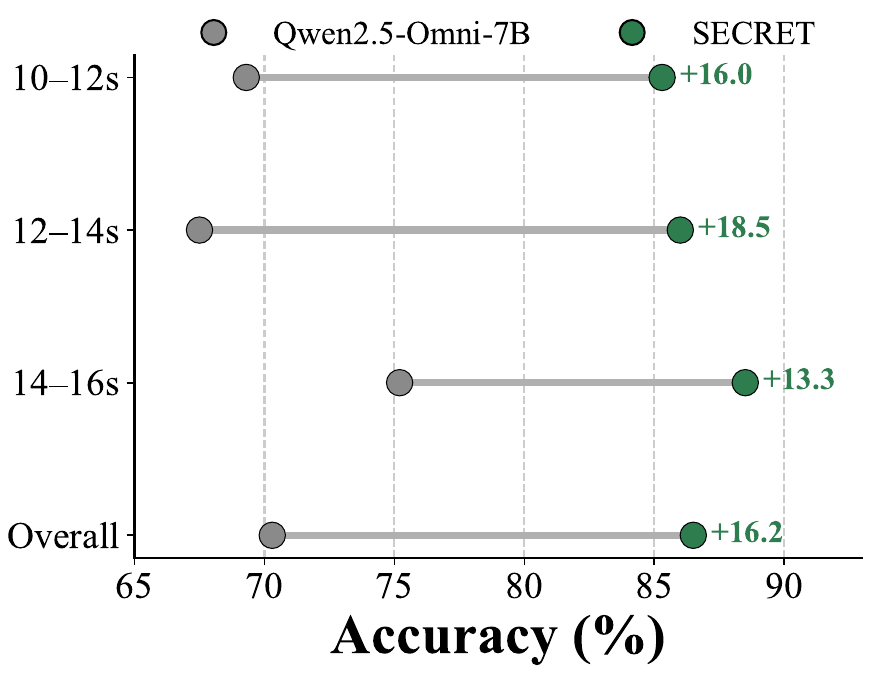}
        \caption{Accuracy across video-duration groups.}
    \end{subfigure}%
    
\vspace{-5pt}

\caption{
\textbf{Fine-grained evaluation on CMM.}
\textbf{(a)} Answer-margin distributions,
where answer-margin is the correct-answer logit minus the incorrect-answer logit.
Diamonds and orange lines denote means and medians; whiskers span min--max.
\textbf{(b)} Accuracy across the reported video-duration groups.
The results show that
\textsc{Secret} maintains substantial gains on longer clips.
}
    \label{fig:exp2}
    \vspace{-5pt}
\end{figure}

\noindent\textbf{RQ2: Which layers are most effective for question
steering, and why?}
Fig.~\ref{fig:ablation}(b) varies $L_1$, the depth up to which
relay interventions are applied before steering.
We test Layers 21--25, which follow the main
modality-to-question information-transfer stage
(see Apdx~\ref{supp:secret_implementation} for the
layer-range selection).
Among the tested depths, VideoLLaMA2-AV-7B performs best
at $L_1=21$ (89.3\%), while Qwen2.5-Omni-7B peaks at
$L_1=25$ (86.4\%).
To investigate this difference, 
Fig.~\ref{fig:ablation}(c) shows that the mean cosine similarity
between positive and negative question representations increases
with depth in VideoLLaMA2-AV-7B but decreases in Qwen2.5-Omni-7B.
For both models, the best-performing depth coincides with
the lowest similarity.
This association suggests that greater positive--negative
separation may provide a more informative steering contrast,
helping explain the different optimal depths.
See Apdx~\ref{supp:pca_analysis} for analysis details
and complementary PCA visualizations.

\noindent\textbf{RQ3: Does \textsc{Secret} generalize to open-ended tasks?}
We evaluate modality-specific captioning under mismatched
audio-video inputs, instructing models to describe only
the requested modality despite receiving both.
Following ACPO~\citep{acpo2026}, \textit{T-CIDEr} measures
target-reference agreement, with higher scores being better.
We additionally report \textit{D-CIDEr}, where lower
distractor-reference overlap suggests less cross-modal leakage,
and \textit{LLM-score}, where higher scores indicate better
target fidelity and less distractor leakage.
Evaluation details are provided in Apdx~\ref{supp:caption_eval}.
As shown in Table~\ref{tab:ablate},
\textsc{Secret} achieves the lowest D-CIDEr and highest
LLM-score while maintaining competitive T-CIDEr across both models and captioning tasks, demonstrating its effectiveness
against source-confused grounding hallucinations
in open-ended generation.

\noindent\textbf{RQ4: How does \textsc{Secret} perform under finer-grained evaluation?}
We examine answer-margin distributions and duration-stratified
accuracy on CMM.
For each example, we compute the answer margin as the
correct-answer logit minus the incorrect-candidate logit,
both evaluated at $X_G$ when predicting the first answer token.
Positive margins favor the correct candidate, while negative
margins favor the incorrect one.
Fig.~\ref{fig:exp2}(a) shows that \textsc{Secret} increases
both mean and median margins relative to the base model
for both backbones, indicating stronger relative support
for correct answers.
Fig.~\ref{fig:exp2}(b) shows consistent accuracy gains
across the most common video-duration groups in CMM.
The gains remain substantial on longer clips, reaching
14.7 percentage points for VideoLLaMA2-AV-7B in the
longest reported group (14--16 seconds).
Together, these analyses provide further insights into the
behavior of \textsc{Secret}.

\section{Related Work}
Multimodal language models (MLLMs) have made substantial progress in integrating and reasoning over heterogeneous inputs~\citep{wang2025athena,li2025chemvlm,wang2025last,zhu2026decoupling}, achieving strong performance across a wide range of multimodal tasks~\citep{zhang2025cross,zhang2024question,ma2026beyond,zhang2025moma,li2026faithful,wang2026self}.
Recent omni-modal models further bring text, audio, and visual information into a unified framework~\citep{xu2025qwen2,Qwen3-Omni,cui2026minicpmo45realtimefullduplex}.
While this integration enables models to use complementary sensory cues, reliable responses require grounding in the source specified by the instruction~\citep{jung2026avcd,zhang2026mitigating,sung2024avhbench}, even when other modalities provide misleading evidence~\citep{chung2026mad,chaubey2026mod}.
This requirement motivates efforts to mitigate source-confused grounding hallucinations and mechanistic studies of how models use multimodal evidence internally.

\noindent\textbf{Source-confused grounding hallucination.}
Prior work identifies source-confused grounding hallucination
in AVLLMs, where cues from one modality induce unsupported
predictions about another~\citep{chung2026mad,chaubey2026mod}.
AVHBench and CMM evaluate these failures across different
directions of cross-modal
interference~\citep{sung2024avhbench,leng2024curse}.
Existing mitigation methods mainly rely on inference-time
correction or training-time alignment.
Inference-time adaptive decoding regulates modality
contributions according to modality dominance, task relevance,
or predictive conflict~\citep{jung2026avcd,chung2026mad,leng2024mitigating}.
Preference-based alignment approaches use multimodal preference pairs
and modality-aware objectives to strengthen grounding in
sensory evidence and reduce inappropriate cross-modal
reliance~\citep{chen2026omnidpo,chaubey2026mod,acpo2026}.
Despite their effectiveness, the internal cross-modal interactions underlying this failure remain insufficiently
understood.
In this work, we identify question states as an internal
relay for cross-modal interference and propose
\textsc{Secret} to steer them toward required-modality
evidence to mitigate hallucination.

\noindent\textbf{Mechanistic understanding of multimodal information utilization.}
Mechanistic studies examine how models integrate and use
multimodal information, providing insights that guide
method design~\citep{nikankin2025same,kim2025map,tong2026flowcut}.
A commonly used approach characterizes modality reliance
at generation positions through attention analyses and
pathway interventions~\citep{selvakumar2026really,yu2026causally}.
These analyses inform attention modulation and contrastive
decoding for hallucination mitigation~\citep{jiang2025devils,jung2026probing}.
Recent studies examine how modality information is integrated
into preceding instruction positions to support
subsequent predictions~\citep{cross_modal,zhang2026instructionanchor,suharitdamrong2026senses}.
Instruction Anchor improves modality following in
vision-language models by identifying and amplifying
attention heads involved in modality
arbitration~\citep{zhang2026instructionanchor}.
Other work also uses these insights to guide token pruning
for more efficient inference in AVLLMs~\citep{suharitdamrong2026senses}.
Beyond these studies, we investigate how non-required
audio-visual cues propagate through the question relay
and lead to source-confused grounding hallucinations
in AVLLMs.
Guided by this diagnosis, we introduce \textsc{Secret},
which uses source-conditioned representation contrasts
to steer question states toward required-source evidence
and mitigate this interference.
\section{Conclusion}

In this work, we investigated source-confused grounding
hallucination in AVLLMs.
Our path-intervention and representation analyses  
reveal a \emph{question-relay} mechanism:
question states
relay interfering cues alongside required-source evidence, undermining grounding in
required-modality evidence.
Building on these findings, we proposed \textsc{Secret},
a training-free method that contrasts question representations
elicited through source-conditioned pathway interventions
to steer generation toward required-modality evidence.
Experiments across three AVLLMs demonstrate consistent
improvements on CMM and AVHBench, while modality-specific
captioning evaluations show improved source grounding in open-ended generation.

\subsection*{AI use statement}

This work investigates source-confused grounding hallucinations in
audio-visual large language models, with experiments on
VideoLLaMA2-AV~\citep{damonlpsg2024videollama2},
Qwen2.5-Omni-7B~\citep{xu2025qwen2}, and
Qwen3-Omni-30B-A3B~\citep{Qwen3-Omni}.
As part of our research pipeline, we use LLMs to identify the
instruction-required modality and extract target objects from
textual questions. We also use GPT-4.1 to evaluate modality-specific
captions for target fidelity and distractor leakage, following the
scoring protocol in Apdx~\ref{supp:gpt_score}.
Additionally, 
we use generative AI tools for
grammatical refinement, and linguistic polishing of the manuscript.
The authors take responsibility for the final content of this work,
including all text, claims, and artifacts produced with AI assistance.
\subsection*{Ethics statement}
This work aims to improve the reliability of audio-visual large language
models by mitigating answers grounded in the wrong modality.
Our evaluation uses existing AVHBench and CMM benchmarks and
modality-specific captioning inputs described in
Apdx~\ref{supp:caption_eval}.
While these insights highlight potential vulnerabilities where safety filters might be bypassed, they primarily establish a structural foundation for developing more robust
and transparent AI safeguards.

\subsection*{Reproducibility statement}

We document the method and evaluation protocols to support reproducibility.
Section~\ref{sec:method} specifies the source-conditioned attention
interventions and question-state update used by \textsc{Secret}.
Apdx~\ref{supp:diagnostics} describes the diagnostic data,
attention-path analyses, and target-object extraction and scoring.
Apdx~\ref{supp:implementation_detail} details model-specific steering
depths, decoding settings, baselines, and intervention variants,
while Apdx~\ref{supp:pca_analysis} describes the representation
analyses. For modality-specific captioning,
Apdx~\ref{supp:caption_eval} provides the evaluation data construction,
generation prompts, text preprocessing, and metric definitions,
including the GPT-4.1 judge prompt and scoring procedure.

\bibliography{iclr2027_conference}
\bibliographystyle{iclr2027_conference}
\appendix
\maketitlesupplementary
Our supplementary materials are summarized as follows:
\begin{itemize}
    \item Appendix~\ref{supp:discussion}:
    Discussion, Limitations, and Future Directions.
    \item Appendix~\ref{supp:diagnostics}:
    Diagnostic Protocols and Attention-Path Robustness Analysis.
    \item Appendix~\ref{supp:implementation_detail}:
    Implementation Details, Baselines, and Intervention Variants.
    \item Appendix~\ref{supp:pca_analysis}:
    Layer-wise Question Representation Analyses.
    \item Appendix~\ref{supp:caption_eval}:
    Modality-Specific Captioning Evaluation.
\end{itemize}
\section{Discussion and Limitation}
\label{supp:discussion}
In this work, we find that question states provide a
shared relay for audio-visual evidence, but can also carry interfering
cues into answer generation. This dual role suggests that effective
multimodal integration requires sensitivity to both the semantic
relevance and the source of incoming information. A cue can be closely
related to the question while still being inappropriate evidence for
the requested modality.
\textsc{Secret} translates this perspective into inference-time control
at the question relay. Its use of source-conditioned representation
contrasts illustrates how mechanistic analysis can inform the design
of training-free interventions while retaining the complete audio-visual
input. More broadly, this connection motivates studying how intermediate
textual states regulate which evidence supports generation.

A promising direction is to disentangle modality-specific cues within question representations, potentially helping models mitigate source-confused grounding hallucinations and make more effective use of information from each modality.
Besides, 
our analysis focuses on cross-modal information flow at the pathway level. Finer-grained analyses, such as examining the roles of individual attention heads, could further clarify the mechanisms underlying source-confused grounding hallucinations.

\section{Diagnostic Protocols}
\label{supp:diagnostics}

\subsection{Dataset Details and Statistics}
\label{supp:data}

We use two subsets of AVHBench~\citep{sung2024avhbench}.
The \emph{Video-Driven Audio Hallucination} subset contains
2,290 questions requiring audio-grounded answers, while the
\emph{Audio-Driven Video Hallucination} subset contains
1,136 questions requiring video-grounded answers.
Together, these subsets comprise 3,426 question--answer
pairs covering both directions of cross-modal interference.
AVHBench draws on VALOR and AudioCaps, covering everyday
scenarios involving people, animals, machinery, and nature.
Its construction distinguishes visible sound sources,
visible but silent objects, and audible sources outside
the camera view.
The latter two categories provide negative examples for
the two hallucination tasks, making these subsets
particularly relevant to studying source-confused grounding hallucination.

\subsection{Required-Modality Identification}
\label{supp:classify}

For \textsc{Observation~1} (\S\ref{sec:intent}), we prompt
Qwen2.5-Omni-7B using only the question text, without audio
or video inputs, and use greedy decoding.
The reference label is \texttt{AUDIO} for Video-Driven Audio
Hallucination and \texttt{VIDEO} for Audio-Driven Video
Hallucination.
The classifier can additionally return \texttt{AMBIGUOUS}
when it cannot identify a unique required modality.
\begin{tcolorbox}[
    breakable,
    colback=gray!5!white,
    colframe=black!75!white,
    title={Required-modality identification prompt},
    fonttitle=\bfseries,
    fontupper=\small
]
Identify the evidence source explicitly requested by the question.
Do not answer the question or infer the source from object names alone.

Return AUDIO if the question asks about sounds or audible events.
Return VIDEO if it asks about visible objects, actions, or events.
Return AMBIGUOUS if neither source is uniquely specified or both
sources are required.

Question: \texttt{\{question\}}

Return exactly one label: AUDIO, VIDEO, or AMBIGUOUS.
Do not include an explanation.
\end{tcolorbox}

\paragraph{Parsing and scoring.}
We remove surrounding whitespace, convert the response to
uppercase, and accept only an exact match to one of the three
labels. All other responses are treated as invalid.
Accuracy is the percentage of evaluated questions whose predicted
label matches the reference label.
Because every question in these two subsets specifies a single
required modality, \texttt{AMBIGUOUS}, invalid outputs, and
incorrect modality labels all count as errors; no such cases
are excluded from the denominator.

\paragraph{Use in \textsc{Secret}.}
The labels \texttt{AUDIO} and \texttt{VIDEO} map to
$\hat r=A$ and $\hat r=V$, respectively, to determine the
positive and negative attention masks.
For an \texttt{AMBIGUOUS} prediction, we randomly select
$\hat r\in\{A,V\}$ before constructing the masks.

\subsection{Details and more experiments for attention-path cutting analysis}
\label{supp:route_cutting}
In this section, we provide more detail for metric and Robustness analysis across different attention-cutting window sizes for attention-path cutting analysis (\S\ref{sec:routing}).

\paragraph{Metric Computation Details.}
For each example $i$ whose original prediction is
source-faithful, let $p_i^{\mathrm{orig}}$ and
$p_{i,\ell}^{\mathrm{cut}}$ denote the target-answer
probabilities before and after cutting a given pathway
within $\mathcal{W}_{\ell}$, respectively.
We compute the relative probability change for each sample
and report the average over the $N$ evaluated samples as a percentage:
\begin{equation}
\Delta P(\ell)
=
\frac{100}{N}
\sum_{i=1}^{N}
\frac{
p_{i,\ell}^{\mathrm{cut}}-p_i^{\mathrm{orig}}
}{
p_i^{\mathrm{orig}}
}.
\label{eq:path_probability_change}
\end{equation}
More negative values indicate that cutting the pathway reduces
target-answer probability, while positive values indicate
an increase.

\begin{figure}[h]
    \centering
    \begin{subfigure}[t]{0.495\linewidth}
        \centering
        \includegraphics[width=0.495\linewidth]
        {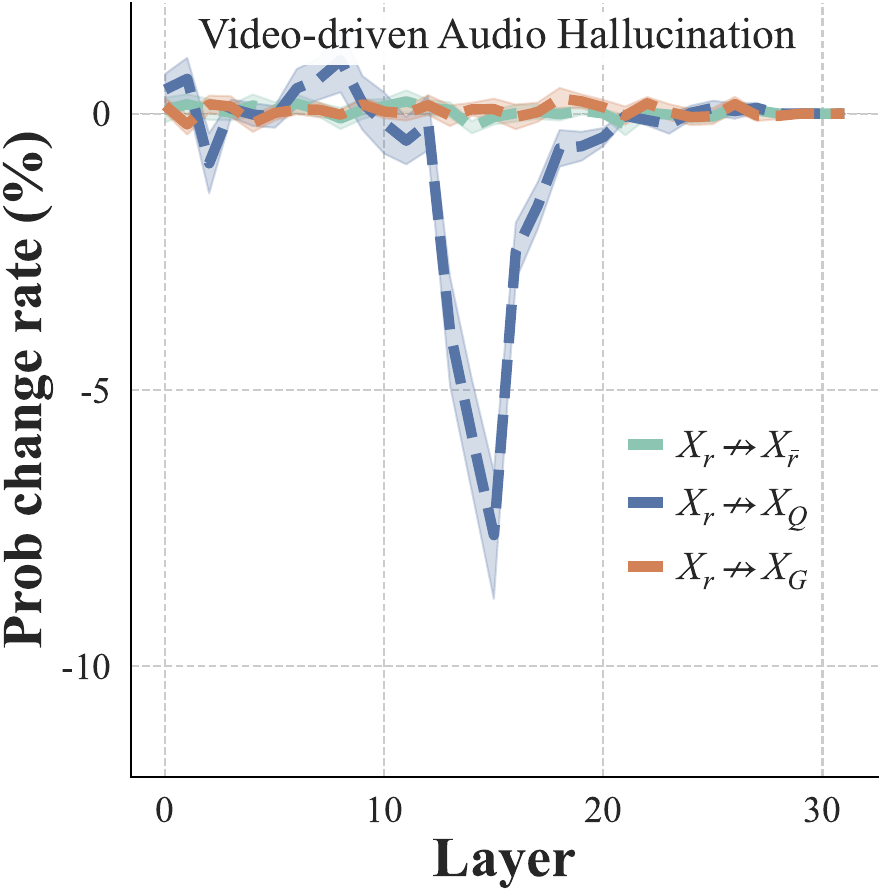}%
        \hfill
        \includegraphics[width=0.495\linewidth]
        {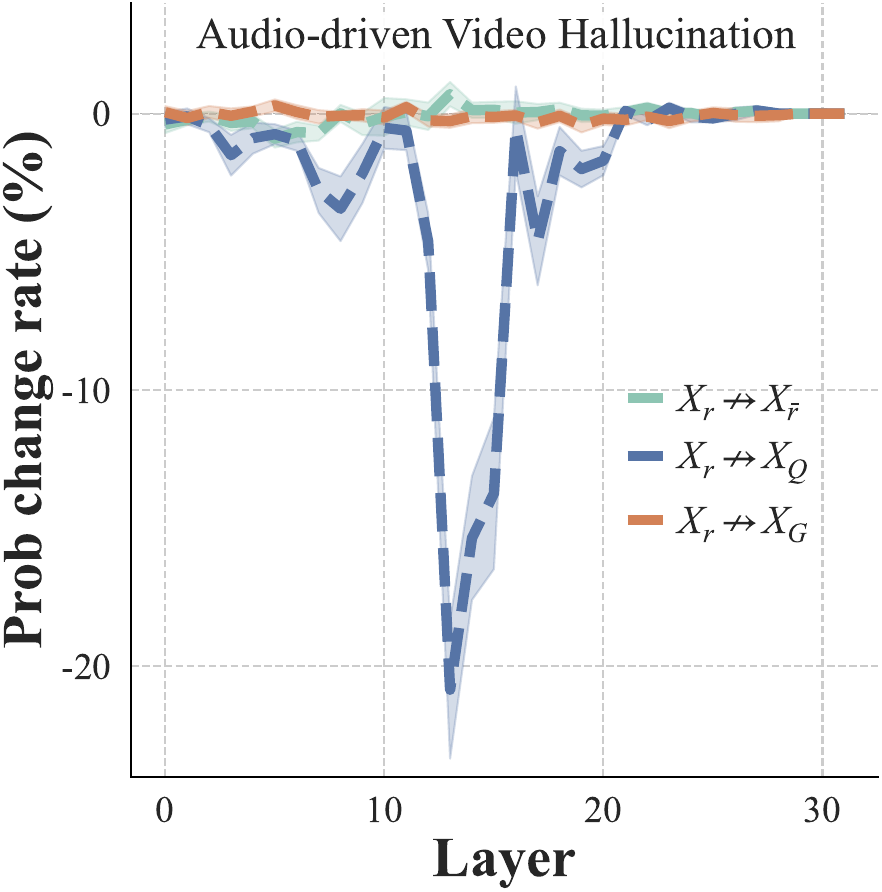}
        \caption{Window size $3$.}
    \end{subfigure}\hfill
    \begin{subfigure}[t]{0.495\linewidth}
        \centering
        \includegraphics[width=0.495\linewidth]
        {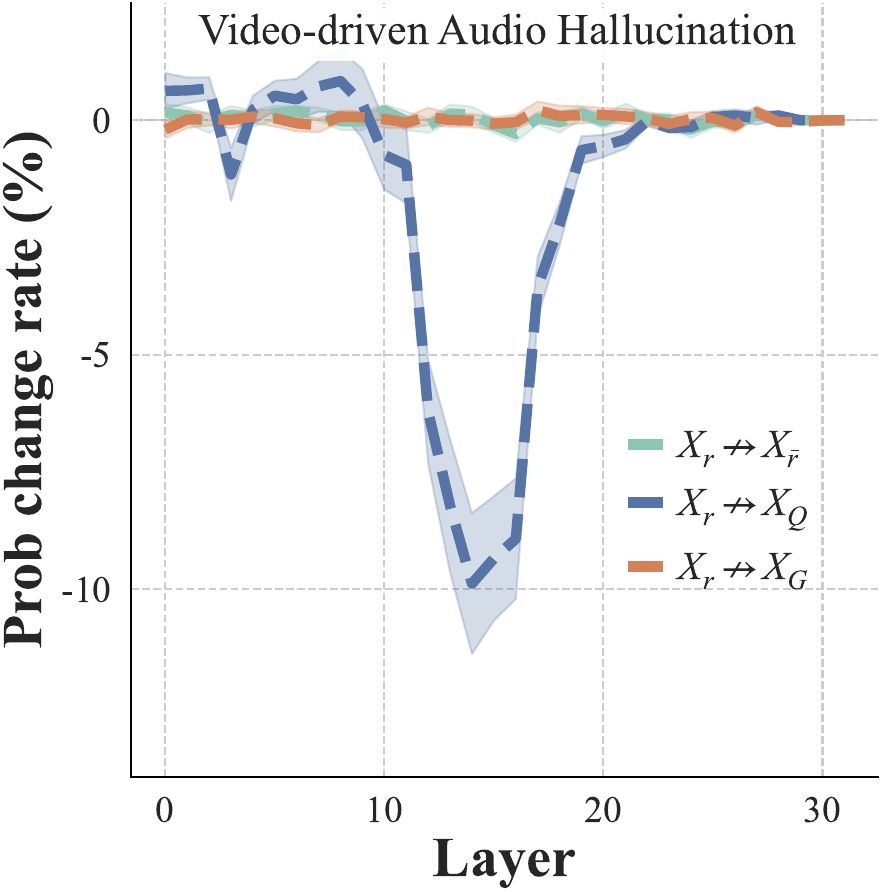}%
        \hfill
        \includegraphics[width=0.495\linewidth]
        {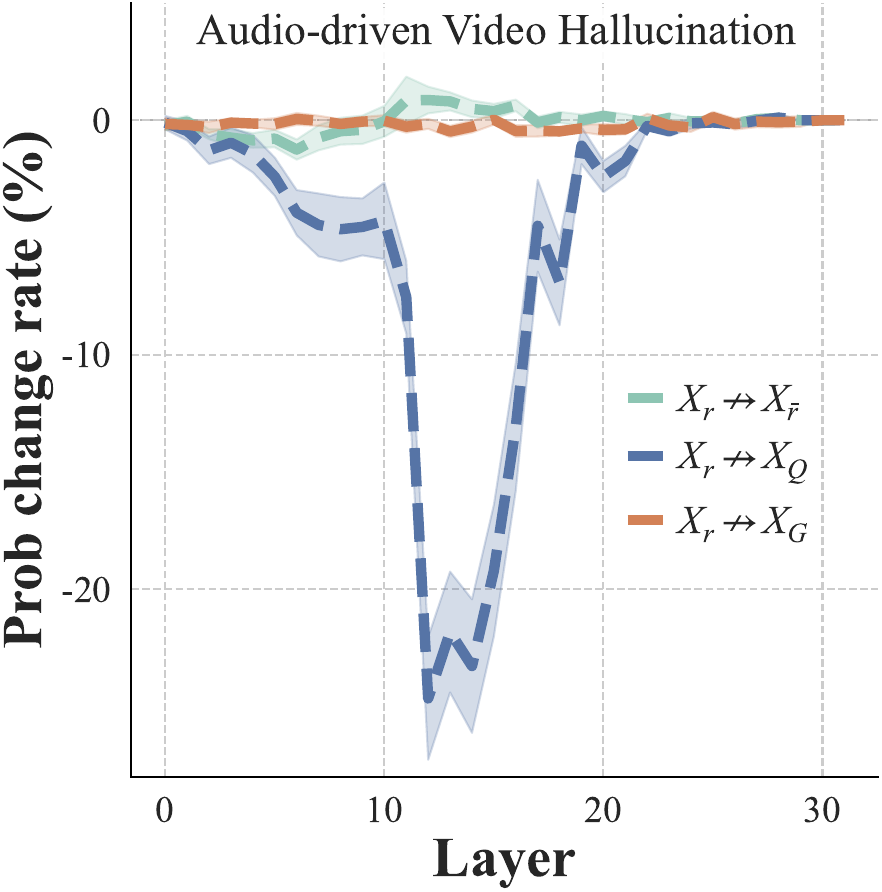}
        \caption{Window size $5$.}
    \end{subfigure}

    \caption{
    \textbf{Robustness to attention-cutting window size in Qwen2.5-Omni-7B.}
    Layer-wise changes in target-answer probability for
    source-faithful examples in Qwen2.5-Omni-7B.
    The horizontal axis indicates the window center.
    Across both settings and window sizes, cutting
    $X_r\!\rightarrow\!X_Q$ produces the most pronounced
    reduction in the middle layers.
    }
    \label{fig:window_robustness}
\end{figure}

\begin{figure}[h]
    \centering
    \begin{subfigure}[t]{0.495\linewidth}
        \centering
        \includegraphics[width=0.98\linewidth]
        {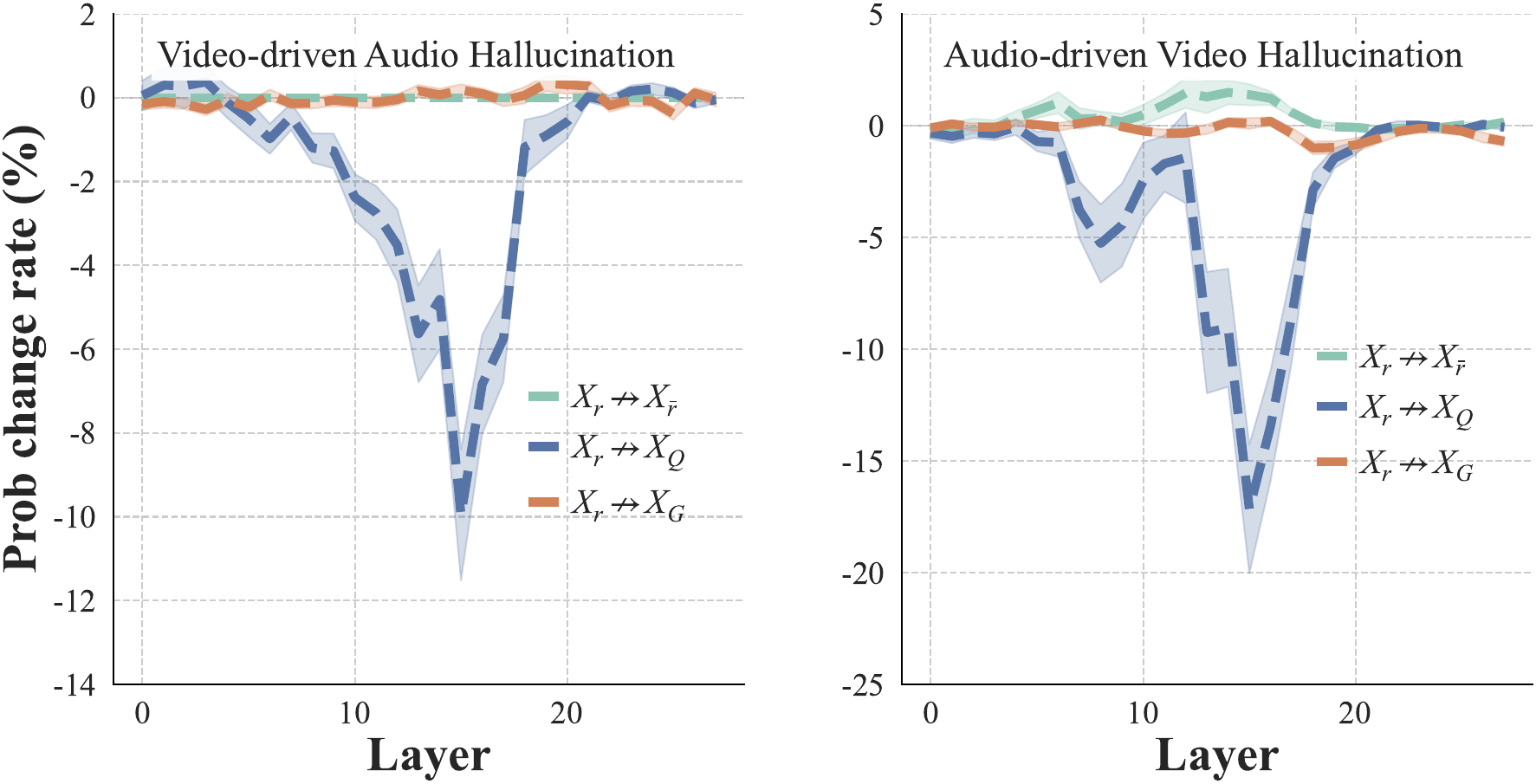}%
        \caption{Window size $3$.}
    \end{subfigure}\hfill
    \begin{subfigure}[t]{0.495\linewidth}
        \centering
        \includegraphics[width=0.98\linewidth]
        {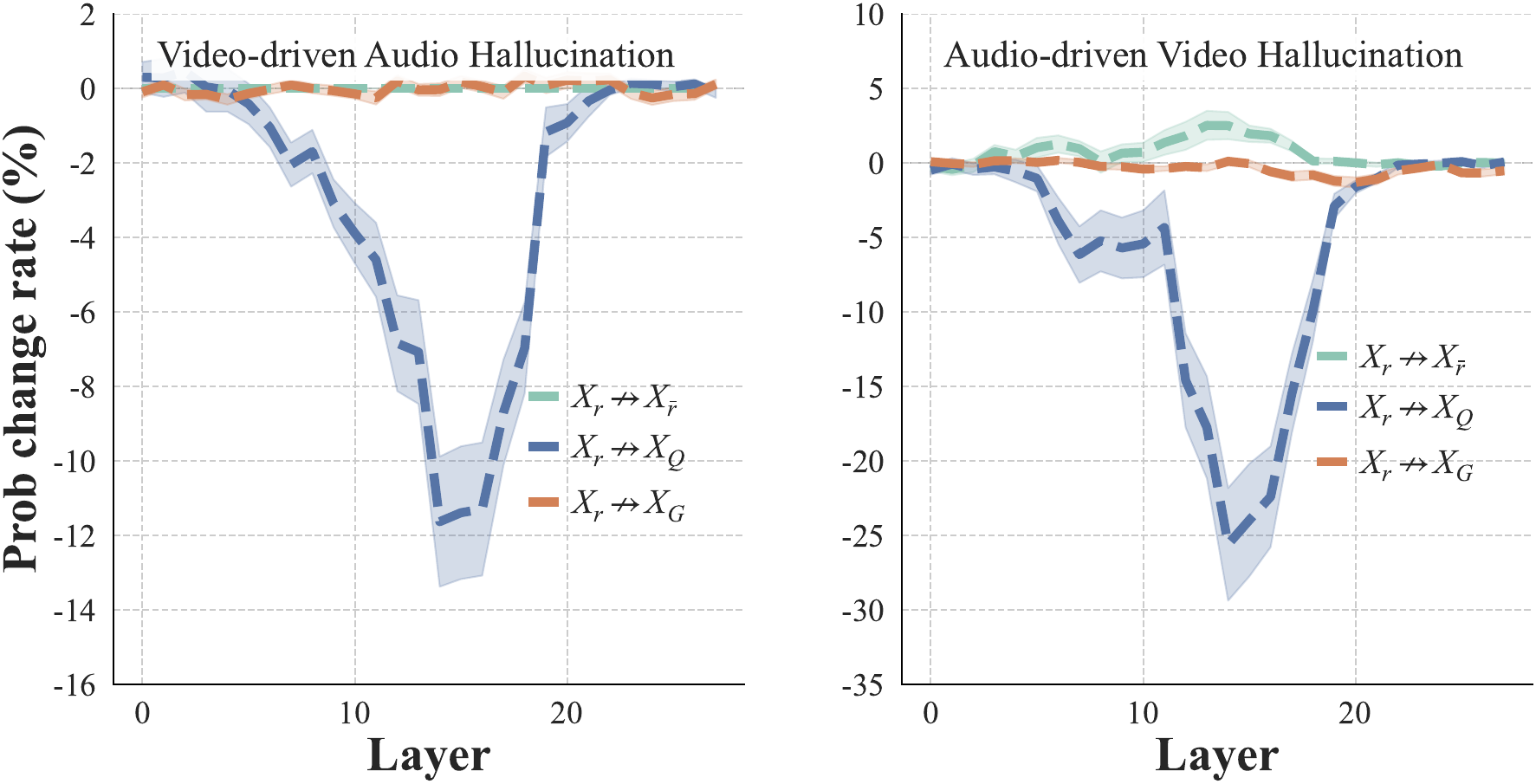}%
        \caption{Window size $5$.}
    \end{subfigure}

    \caption{
    \textbf{Robustness to attention-cutting window size for VideoLLaMA-AV-7B.}
    Layer-wise changes in target-answer probability for
    source-faithful examples in VideoLLaMA-AV-7B.
    The horizontal axis indicates the window center.
    Across both settings and window sizes, cutting
    $X_r\!\rightarrow\!X_Q$ produces the most pronounced
    reduction in the middle layers.
    }
    \label{fig:model_robot}
\end{figure}

\paragraph{Robustness analysis across different AVLLMs and attention-cutting window sizes.}
The main analysis in Fig.~\ref{fig:finding1_2} uses a
seven-layer window.
To assess whether its conclusions depend on this choice,
we repeat the pathway analysis with three- and five-layer
windows, comparing the same three pathways in both
hallucination settings.

As shown in Fig.~\ref{fig:window_robustness} and Fig.~\ref{fig:model_robot}, cutting
$X_r\!\rightarrow\!X_Q$ produces substantially larger
reductions in target-answer probability in the middle
layers than cutting $X_r\!\rightarrow\!X_G$ or
$X_r\!\rightarrow\!X_{\bar r}$.
Changing the window size affects the magnitude and
layer-wise extent of the reductions, but the strongest
effects remain concentrated in the middle layers and
associated with the modality-to-question pathway.
Together, these findings
support the role of question states as an important relay
for required-source evidence across the tested window sizes.

\subsection{Object Extraction and Target-Object Scores}
\label{supp:object_evidence}

\paragraph{Target-object extraction.}
An LLM parser (Qwen3-32B-Instruct) identifies the target object from each question
without accessing the audio or video.
For example, ``Do you hear piano music?'' yields ``piano''.
The following prompt specifies the extraction task.

\begin{tcolorbox}[
    breakable,
    colback=gray!5!white,
    colframe=black!75!white,
    title={Target-object extraction prompt},
    fonttitle=\bfseries,
    fontupper=\small,
    before upper={
        \setlength{\parindent}{0pt}
        \setlength{\parskip}{4pt}
    }
]
Extract the target object explicitly queried in the question.
Return the shortest noun phrase that preserves its identity.
Do not answer the question or infer objects that are not mentioned.
If no unique target object can be identified, return NONE.

Example:\\
Question: Do you hear piano music?\\
Output: piano

Question: \texttt{\{question\}}

Return only the object name or NONE, without explanation.
\end{tcolorbox}

\paragraph{Target-object score.}
We use Logit Lens~\citep{geva2022transformer} to quantify
target-object signals in question states.
For a given example, let $\mathbf{h}_j^\ell$ denote the
hidden state at question position $j$ and layer $\ell$.
For the extracted object $o$, let $\mathbf{w}_o$ denote
the output-head weight vector corresponding to its
representative vocabulary token.
We compute the object's logit at each question position
and take the maximum across these
positions:
\begin{equation}
\begin{aligned}
S^\ell(o) &= \max_{j \in {X}_Q} z_j^\ell(o), \quad
z_j^\ell(o) &= \mathbf{w}_o^\top \mathbf{h}_j^\ell
\end{aligned}
\label{eq:target_object_score}
\end{equation}
where $X_Q$ is the set of question-token positions.
This yields one target-object score per example and layer.
A higher score indicates stronger target-object support
within question states.

\paragraph{Comparison and aggregation.}
We compute the score separately for \textit{Original}
and \textit{Intervened}, using the same target object.
And
the two runs retain identical inputs. 
At each layer, Fig.~\ref{fig:finding3}(a, left) reports
the percentage of analyzed examples whose score is
strictly lower in \textit{Intervened} than in
\textit{Original}.
The right panel reports the mean score across examples
for each run.

\section{Experimental and Implementation Details}
\label{supp:implementation_detail}

\subsection{Implementation details of \textsc{Secret}}
\label{supp:secret_implementation}

\begin{figure}[htbp]
    \centering
    \includegraphics[width=\linewidth]{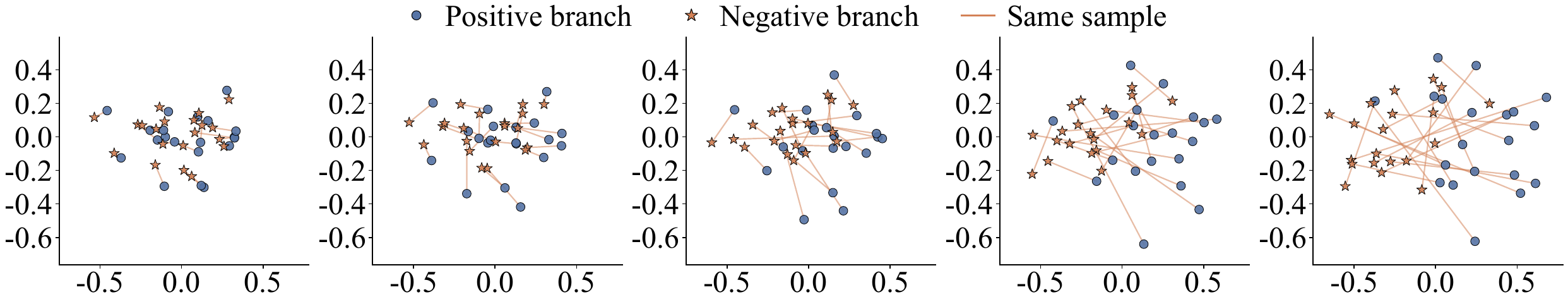}
    \caption{PCA visualization of question representations
    in Qwen2.5-Omni-7B across layers 21--25 (from left to right).
    Blue circles and orange stars denote positive and
    negative representations, respectively;
    lines connect paired representations.}
    \label{fig:pca-qwen}
\end{figure}

\begin{figure}[htbp]
    \centering
    \includegraphics[width=\linewidth]{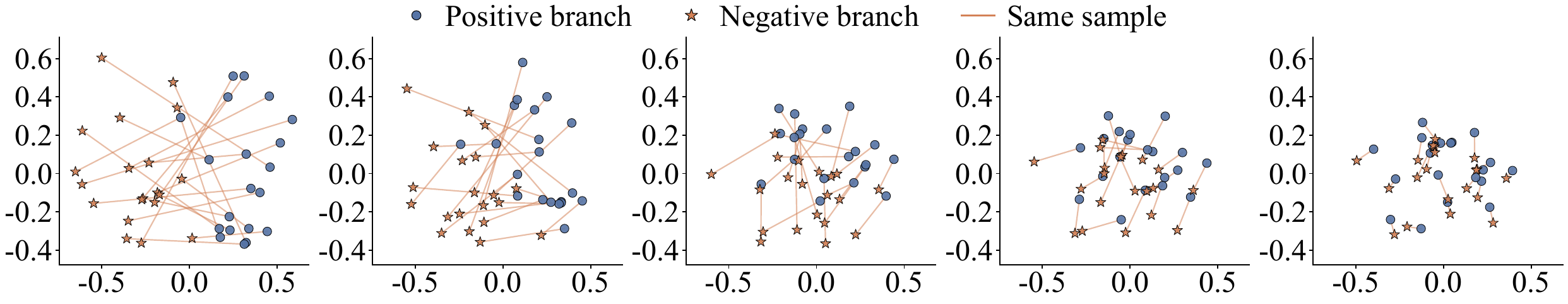}
    \caption{PCA visualization of question representations
    in VideoLLaMA2-AV-7B across layers 21--25 (from left to right). Blue circles and orange stars denote positive and
    negative representations, respectively;
    lines connect paired representations.}
    \label{fig:pca-videollama}
\end{figure}

We use greedy decoding for all AVLLMs and select
model-specific steering depths $L_1$ for \textsc{Secret}.
Steering is applied after the main modality-to-question
information-transfer stage.
For Qwen2.5-Omni-7B, the routing analysis in
Fig.~\ref{fig:finding1_2} motivates candidate depths
after Layer 20.
Our visualizations indicate a similar range for
VideoLLaMA2-AV-7B, while that for Qwen3-Omni-30B-A3B
starts at Layer 28.
Guided by the separation between positive and negative
question representations in Figs.~\ref{fig:pca-qwen} and~\ref{fig:pca-videollama}, we set $L_1=25$, $21$, and $32$
for Qwen2.5-Omni-7B, VideoLLaMA2-AV-7B, and
Qwen3-Omni-30B-A3B, respectively.

At the selected depth, we retain the positive, negative,
and original question-token states for steering,
excluding the generation position $X_G$.
After aligning these states by token position, we apply
the update in Eq.~\ref{eq:question_steering}.
The updated question states are combined with the remaining
token states from the original branch and passed through
the remaining Transformer layers.

\subsection{Baselines and Intervention Variants}
\label{supp:baselines}

\paragraph{Baselines.}
For the four-branch VCD extension, the three contrastive
branches modify video only, audio only, and both modalities.
When implemented through modality removal, these branches
receive audio--question, video--question, and question-only
inputs, respectively~\citep{chung2026mad}.
MAD~\citep{chung2026mad} constructs full audio-visual,
video-only, audio-only, and question-only branches
by omitting the corresponding modality inputs.
The question remains unchanged, and all branches receive
the same generated answer prefix at each decoding step.
AVCD~\citep{jung2026avcd} implements attentive masking
by setting selected token representations to zero
while retaining their sequence positions.
Its contrastive branches mask different combinations
of the less dominant modalities.

\paragraph{Intervention variants.}
The variants in RQ1 (\S\ref{sec:analysis}) and Table~\ref{tab:ablate} modify
how the steering direction is constructed or applied.
\textit{Gen-Interv.} redirects the positive and negative
pathway cuts from $X_Q$ to $X_G$, while retaining the
full audio-visual input.
The resulting contrast is used to steer the original state
at $X_G$ during prefill.
The positive and negative branches redirect the corresponding
pathway cuts from question positions to the generation position.
\textit{Removal-Interv.} constructs positive and negative representations
by removing modality inputs instead of cutting internal
attention pathways.
The positive branch retains
the required modality only,
while the negative branch retains
the interfering modality only.
Both branches preserve the question text.
Because modality removal can change token positions,
the states used to construct the contrast are aligned by
question-token order.
The resulting difference is applied to
the original question states in the
full-input branch.

\section{Representation Analyses}

\label{supp:pca_analysis}

For each model, we compare positive and negative question
representations before steering, using the same CMM
examples across all tested depths.

\paragraph{Cosine similarity.}
We compute cosine similarity between positive and negative
representations at matching question-token positions,
then average these similarities over all question tokens
in the analyzed examples.
Figure~\ref{fig:ablation}(c) reports the resulting
layer-wise mean similarities.

\paragraph{PCA visualization.}
We further visualize the question representations using
two-dimensional PCA.
At each layer, PCA is fitted jointly to the positive and
negative representations, with lines connecting paired
representations.
These projections provide a qualitative view of their
separation; quantitative comparisons across layers rely
on cosine similarity in the original representation space.

\section{Modality-Specific Captioning Evaluation}
\label{supp:caption_eval}

\subsection{Data, Generation, and Preprocessing}

Following ACPO~\citep{acpo2026}, we evaluate audio-target and
video-target captioning on 400 audio-swapped examples each.
Models receive both modalities but describe only the requested one.
ACPO constructs its evaluation set from AVHBench captioning clips
with diverse and distinct audio events. It retains each video's
visual content and replaces its audio with a track from another clip,
using an LLM to rank candidate tracks and select plausible mismatches.
We use the modality-specific reference captions associated with
these inputs.

All three metrics are computed on swapped inputs; original aligned
inputs are excluded because semantic overlap makes distractor leakage
harder to distinguish from valid target content.
For video $v_A$ paired with audio $a_B$, audio-target captioning uses
$a_B$'s audio caption as the target reference and $v_A$'s visual caption
as the distractor reference; video-target captioning reverses these roles.
Each example has one reference per modality.
Scores are computed separately for each model, method, and captioning
target.

We adopt the generation prompts from ACPO:
``Describe what you hear.'' for audio-target captioning and
``Describe what you see.'' for video-target captioning.
We use the same greedy decoding and model-specific steering depths
as in Apdx~\ref{supp:secret_implementation}.
All methods use the same text preprocessing: truncate at the first
\texttt{Human:} or \texttt{User:}, retain the first complete sentence,
and strip surrounding whitespace.
No paraphrasing or within-sentence content removal is applied.
The processed caption is used for all metrics and human validation.

\subsection{Target and Distractor CIDEr}

Let $\mathcal C$, $\mathcal T$, and $\mathcal D$ be the generated,
target-reference, and distractor-reference corpora, paired by sample ID.
Following ACPO~\citep{acpo2026} for target-reference evaluation, we report
\begin{align}
    \mathrm{T\mbox{-}CIDEr}
    &=100\times\mathrm{CIDEr}(\mathcal C,\mathcal T),
    \label{eq:target_cider}\\
    \mathrm{D\mbox{-}CIDEr}
    &=100\times\mathrm{CIDEr}(\mathcal C,\mathcal D).
    \label{eq:distractor_cider}
\end{align}
We use standard \texttt{pycocoevalcap} implementation,\footnote{
\url{https://github.com/salaniz/pycocoevalcap/tree/master/cider}}
with TF--IDF-weighted 1--4-grams and the default length penalty
$\sigma=6$. 
The two calls use identical predictions and sample IDs and differ only
in the reference corpus. Higher T-CIDEr indicates stronger target-reference
agreement; lower D-CIDEr indicates less distractor-reference overlap.
Low D-CIDEr alone is insufficient, since empty or generic descriptions
 also avoid distractor content.

\subsection{LLM-Based Grounding Evaluation}
\label{supp:gpt_score}

The judge receives the target modality, both references, and the generated
caption, rather than raw audio/video. It assigns target quality
$Q_i\in\{1,\ldots,5\}$ and distractor leakage $L_i\in\{0,\ldots,3\}$
using the prompt below. Shared reference content is not counted as
leakage; unrelated hallucinations reduce target quality.
The sample score and reported average are
\begin{align}
    G_i&=\max(1,Q_i-L_i),\label{eq:gpt_grounding_sample}\\
    \mathrm{LLM\mbox{-}score}
    &=\frac{1}{N}\sum_{i=1}^{N}G_i.\label{eq:gpt_grounding_mean}
\end{align}
The resulting LLM-score ranges from 1 to 5,
with higher scores indicating better modality grounding.
This score jointly assesses target fidelity and distractor leakage;
it is not calculated from numerical CIDEr scores.
The lower bound is applied before averaging, and an empty caption
receives a score of 1 through its target-quality score.

We use GPT-4.1 with \texttt{temperature=0}, hide model and method
names, and evaluate each caption independently.
The sample score is recomputed in code from the returned $Q_i$ and $L_i$.

\paragraph{Judge prompt.}
The fields in braces are replaced with the requested modality, references,
and preprocessed prediction for each sample.

\begin{tcolorbox}[
    breakable,
    colback=gray!5!white,
    colframe=black!75!white,
    title={GPT modality-grounding evaluation prompt},
    title after break={GPT modality-grounding evaluation prompt (continued)},
    fonttitle=\bfseries,
    fontupper=\small,
    before upper={\setlength{\parindent}{0pt}\setlength{\parskip}{4pt}}
]
You are evaluating whether a generated caption follows the requested
audio or video modality under deliberately mismatched audio-video input.

Target modality: \texttt{\{AUDIO or VIDEO\}}

Target reference:\\
\texttt{\{target\_caption\}}

Distractor reference from the other modality:\\
\texttt{\{distractor\_caption\}}

Generated caption:\\
\texttt{\{prediction\}}

Evaluate semantic meaning rather than exact wording.

First assign:\\
1. \texttt{target\_quality}: an integer from 1 to 5\\
2. \texttt{distractor\_leakage}: an integer from 0 to 3

Target quality:\\
5 = complete and accurate target description\\
4 = mostly correct with minor omissions\\
3 = partially correct or overly generic\\
2 = weakly related to the target\\
1 = absent, incorrect, or contradictory target information

Distractor leakage:\\
0 = no distractor-specific information\\
1 = minor or ambiguous distractor influence\\
2 = substantial mixture of target and distractor information\\
3 = distractor dominates the generated caption

Information shared by both references must not be treated as distractor
leakage. Unrelated hallucinations reduce target quality but are not
automatically distractor leakage.

Compute:\\
\texttt{overall\_score = max(1, target\_quality - distractor\_leakage)}

Return JSON only:
\begin{verbatim}
{
  "target_quality": 1,
  "distractor_leakage": 0,
  "overall_score": 1,
  "reason": "Brief explanation within 30 words."
}
\end{verbatim}
\end{tcolorbox}

\subsection{Human Validation of LLM Scores}
\label{supp:human_validation}

\paragraph{Pairwise preferences.}
We sampled caption pairs across both target modalities, both
backbone models, and the evaluated methods.
Each pair consisted of two processed captions generated by different
methods for the same swapped input and target modality.
Annotators received the requested modality, the same target and
distractor references provided to GPT-4.1, and the two captions in
randomized order, with method identities and GPT scores hidden.
Using the same criteria of target accuracy, completeness, and
distractor leakage, they judged the first caption as better,
the second as better, or the two as comparable.
Sampling did not depend on which method won or on GPT's preference.

\paragraph{Pairwise agreement.}
For comparison pair $j$, let $h_j\in\{-1,0,1\}$ denote the human
preference, where $1$ favors the first caption, $-1$ favors the second,
and $0$ denotes a tie.
The GPT preference is derived directly from the existing per-caption
scores in Eq.~\ref{eq:gpt_grounding_sample}:
\begin{equation}
    g_j=\operatorname{sign}\!\left(G_j^{(1)}-G_j^{(2)}\right).
\end{equation}
For $M$ evaluated pairs, we define
\begin{equation}
    \mathrm{Pairwise\ Agreement}
    =\frac{100}{M}\sum_{j=1}^{M}\mathbb{I}[h_j=g_j].
    \label{eq:human_gpt_agreement}
\end{equation}
Agreement requires matching preferences, including when both judges
indicate a tie. A tie from only one judge counts as disagreement,
and all ties remain in the denominator.
This metric assesses whether LLM-score differences reflect human
preferences without requiring matching absolute scores or a separate
pairwise GPT prompt.
The results show that GPT--human agreement is 89\% for audio-target captioning and 87\% for video-target captioning, compared with human--human agreement of 93\% and 90\%, respectively.

\end{document}